\documentclass[onecolumn,draftclsnofoot]{IEEEtran}
\usepackage{cite}
\usepackage{stmaryrd}
\SetSymbolFont{stmry}{bold}{U}{stmry}{m}{n}

\usepackage{bm,mathtools,amsfonts,amssymb,bbm,euscript,setspace,tikz,steinmetz,enumitem,wrapfig,amsthm,amsmath,stmaryrd}

\usepackage{tikz}
\usetikzlibrary{shapes.geometric, arrows.meta, fit, positioning, calc}

\usepackage[normalem]{ulem}
\usepackage[mode=buildnew]{standalone}
\usepackage{mathabx} 
\usepackage[utf8]{inputenc} 
\usepackage[T1]{fontenc}    
\usepackage{url}            
\usepackage{booktabs}       
\usepackage{nicefrac}       
\usepackage{caption,subcaption}

\newtheorem{theorem}{Theorem}
\newtheorem{lemma}[theorem]{Lemma} 
 
\newtheorem{remark}[theorem]{Remark}

\newtheorem{definition}[theorem]{Definition}

\newtheorem{assumption}[theorem]{Assumption}

\usepackage{mleftright}\mleftright

\definecolor{darkgreen}{rgb}{0, 0.5, 0}
\definecolor{midgreen}{rgb}{0, 0.7, 0}

\definecolor{darkred}{RGB}{128, 0, 0}
\definecolor{darkpink}{RGB}{230, 51, 106}
\definecolor{darkestred}{RGB}{139, 0, 0}

\newcommand{\stkout}[1]{\ifmmode\text{\sout{\ensuremath{#1}}}\else\sout{#1}\fi}

\newcommand{\vc}[1]{\mathbf{#1}} 

\newcommand{\eg}{\emph{e.g., }}
\newcommand{\ie}{\emph{i.e., }}

\DeclareMathOperator{\diag}{diag}

\DeclareMathOperator{\gen}{gen}

\DeclareMathOperator{\unif}{Unif}

\DeclareMathOperator{\E}{\mathbb{E}}
\DeclareMathOperator{\R}{\mathbb{R}}

\makeatletter
\DeclareFontFamily{OMX}{MnSymbolE}{}
\DeclareSymbolFont{MnLargeSymbols}{OMX}{MnSymbolE}{m}{n}
\SetSymbolFont{MnLargeSymbols}{bold}{OMX}{MnSymbolE}{b}{n}
\DeclareFontShape{OMX}{MnSymbolE}{m}{n}{
	<-6>  MnSymbolE5
	<6-7>  MnSymbolE6
	<7-8>  MnSymbolE7
	<8-9>  MnSymbolE8
	<9-10> MnSymbolE9
	<10-12> MnSymbolE10
	<12->   MnSymbolE12
}{}
\DeclareFontShape{OMX}{MnSymbolE}{b}{n}{
	<-6>  MnSymbolE-Bold5
	<6-7>  MnSymbolE-Bold6
	<7-8>  MnSymbolE-Bold7
	<8-9>  MnSymbolE-Bold8
	<9-10> MnSymbolE-Bold9
	<10-12> MnSymbolE-Bold10
	<12->   MnSymbolE-Bold12
}{}

\let\llangle\@undefined
\let\rrangle\@undefined
\DeclareMathDelimiter{\llangle}{\mathopen}%
{MnLargeSymbols}{'164}{MnLargeSymbols}{'164}
\DeclareMathDelimiter{\rrangle}{\mathclose}%
{MnLargeSymbols}{'171}{MnLargeSymbols}{'171}
\makeatother

\usepackage[utf8]{inputenc} 
\usepackage[T1]{fontenc}    
\usepackage{hyperref}       
\usepackage{url}            
\usepackage{booktabs}       
\usepackage{amsfonts}       
\usepackage{nicefrac}       
\usepackage{microtype}      
\usepackage{xcolor}         

\begin{document}

\title{Multiview Representation Learning via \\Distributed Joint Latent Space Structuring}

\author{Milad Sefidgaran \qquad Piotr Krasnowski \qquad Abdellatif Zaidi

\thanks{Milad Sefidgaran (email: milad.sefidgaran2@huawei.com), Piotr Krasnowski (email: piotr.g.krasnowski@huawei.com), and Abdellatif Zaidi (email: abdellatif.zaidi@univ-eiffel.fr) are with the Huawei Paris Fourier Research Center, 92100 Boulogne-Billancourt, France. Abdellatif Zaidi is also with the Laboratoire d'Informatique Gaspard Monge, Université Gustave Eiffel, 77420 Champs-sur-Marne, France}

}

\maketitle

\begin{abstract}
We study distributed multiview representation learning, a problem in which $K$ clients each observe a distinct but possibly statistically correlated view. The clients independently extract local representations from their views, which are then used by a central decoder for joint target estimation. One central difficulty is that, since the clients are not allowed to communicate with each other, they must autonomously decide what to encode. We study this coordination problem from a generalization error perspective. For both classification and regression tasks, we derive novel generalization bounds expressed in terms of the Minimum Description Length (MDL) of the joint latent variables across all views and across both training and test datasets. Our structure-aware bound reveals that statistical correlations among the extracted representations tighten the bound, providing theoretical grounding for the empirically observed benefits of cross-view feature alignment. Perhaps counterintuitively, our findings imply that encoders may benefit from extracting redundant representations. Motivated by these bounds, we introduce a data-dependent Gaussian product mixture prior that can be learned and applied in a fully distributed manner. The joint structure of this multiview prior captures inter-view dependencies that are typically discarded by marginal-only approaches. Comprehensive experiments across multiple datasets, encoder architectures, numbers of views, and distortion settings demonstrate the effectiveness of our proposed approach.\footnote{This paper is presented in part at ISIT 2025 \cite{sefidgaran2025multi}.}
\end{abstract}

\section{Introduction} \label{sec:intro}
Understanding when models trained on finite samples remain reliable on unseen data is a central goal of statistical learning theory. This property, formalized through \emph{generalization error}, has been studied through a variety of lenses, including compression-based
approaches~\cite{littlestone1986relating,blumer1987occam,arora2018stronger,%
blum2003pac,suzuki2018spectral,hsu2021generalization,barsbey2021heavy,%
hanneke2019sharp,hanneke2019sample,bousquet2020proper,hanneke2021stable,%
hanneke2020universal,cohen2022learning,Sefidgaran2022,sefidgaran2024data},
information-theoretic
tools~\cite{russozou16,xu2017information,steinke2020reasoning,esposito2020,%
Bu2020,haghifam2021towards,neu2021informationtheoretic,aminian2021exact,%
harutyunyan2021,Zhou2022,lugosi2022generalization,hellstrom2022new,sefidgaran2026tighter},
PAC-Bayes
methods~\cite{seeger2002pac,langford2001not,catoni2003pac,maurer2004note,%
germain2009pac,tolstikhin2013pac,begin2016pac,thiemann2017strongly,%
dziugaite2017computing,neyshabur2018pacbayesian,rivasplata2020pac,%
negrea2020defense,negrea2020it,viallard2021general}, and intrinsic
dimension-based
analyses~\cite{simsekli2020hausdorff,birdal2021intrinsic,hodgkinson2022generalization,lim2022chaotic}.

A widely adopted strategy for improving generalization is the
\emph{encoder-decoder} paradigm, broadly known as \textit{representation
learning}. The encoder maps each input to a compact \emph{representation}
intended to control model complexity, while the decoder minimizes empirical
risk on top of those representations. The information bottleneck (IB),
proposed in~\cite{tishby2000information} and subsequently extended
in~\cite{shamir2010learning,alemi2016deep,estella2018distributed,kolchinsky2019nonlinear,fischer2020conditional,rodriguez2020convex,kleinman2022gacs},
is arguably the most studied instantiation of this idea. The IB principle
treats Shannon's mutual information between the input and the representation
as a surrogate for generalization quality, an assumption that has been repeatedly challenged on both theoretical and empirical
grounds~\cite{kolchinsky2018caveats,rodriguez2019information,amjad2019learning,geiger2019information,dubois2020learning,lyu2023recognizable,sefidgaran2023minimum}.
Encoder-decoder generalization has also been analyzed through PAC-Bayes
lenses~\cite{mbacke2023statistical,cherief2022pac,epstein2019generalization},
yielding bounds on reconstruction or prediction loss in terms of
encoder/decoder parameter complexity; yet these analyses typically do not
single out the role of the latent representations themselves. A more direct
connection was established in~\cite{sefidgaran2023minimum,futami2025information},
which build on~\cite{blum2003pac} to derive PAC-MDL bounds linking
generalization to the \emph{minimum description length} (MDL) of the latent
variable and to the \emph{geometric compression}
of~\cite{geiger2019information}. Building on this line of work,
\cite{sefidgaran2025generalization} introduced a data-dependent Gaussian
mixture prior over the latent distribution and obtained tighter guarantees
for the single-encoder case. The MDL framework has also been adapted to
VQ-VAE architectures in~\cite{futami2025information}, covering the
discrete-latent-variable regime.

\begin{figure}[t]
\centering
\begin{tikzpicture}[
    font=\small,
    block/.style={
        rectangle, rounded corners=3pt, draw=black, thick,
        minimum width=2.4cm, minimum height=0.7cm,
        align=center, fill=white
    },
    arrow/.style={-{Stealth[length=2mm]}, thick},
]

\node (Y) at (0, 0) {$Y$};

\node (X1) at (2.0,  1.4) {$X_1$};
\node (Xd) at (2.0,  0.0) {$\vdots$};
\node (XK) at (2.0, -1.4) {$X_K$};

\node[block, fill=green!8] (E1) at (4.5,  1.4) {$P_{U_1|X_1,W_{e,1}}$};
\node (Ed) at (4.5,  0.0) {$\vdots$};
\node[block, fill=green!8] (EK) at (4.5, -1.4) {$P_{U_K|X_K,W_{e,K}}$};

\node (U1) at (7.0,  1.4) {$U_1$};
\node (Ud) at (7.0,  0.0) {$\vdots$};
\node (UK) at (7.0, -1.4) {$U_K$};

\node[block, fill=blue!8] (Dec) at (9.0, 0) {$P_{\hat{Y}|\mathbf{U},W_d}$};

\node (Yhat) at (11.0, 0) {$\hat{Y}$};

\draw[draw=gray!60, thick, dashed, rounded corners=5pt]
    (1.2, 1.9) rectangle (7.7, 0.9);
\node[font=\footnotesize, text=gray] at (1.8, 1.75) {Client $1$};

\draw[draw=gray!60, thick, dashed, rounded corners=5pt]
    (1.2, -0.9) rectangle (7.7, -1.9);
\node[font=\footnotesize, text=gray] at (1.8, -1.7) {Client $K$};

\draw[draw=blue!40, thick, rounded corners=5pt]
    (7.5, -0.6) rectangle (10.5, 0.7);
\node[font=\footnotesize, text=blue!60] at (9.0, 0.55) {Server};

\draw[draw=red!60, thick, dashed] (4, 0.85) -- (4, -.85);
\node[font=\footnotesize, text=red!70, rotate=90] at (3.8, 0) {no comm.};


\draw[arrow] (Y) -- (X1);
\draw[arrow] (Y) -- (XK);

\draw[arrow] (X1) -- (E1);
\draw[arrow] (XK) -- (EK);

\draw[arrow] (E1) -- (U1);
\draw[arrow] (EK) -- (UK);

\draw[arrow] (U1) -- (Dec);
\draw[arrow] (UK) -- (Dec);
\draw[arrow] (Ud) -- (Dec);

\draw[arrow] (Dec) -- (Yhat);

\end{tikzpicture}
\caption{Distributed multiview representation learning setup.}
\label{fig:distributed}
\vspace{-0.5 cm}
\end{figure}
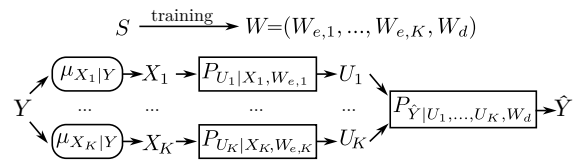

The encoder-decoder architecture described above, comprising a single
encoder feeding a single decoder, is often called \textit{centralized}
representation learning, since all training data reside at one location.
Many practical deployments break this assumption: sensor arrays, multi-modal
pipelines, and split inference systems each route a distinct data stream
to its own processing node, which must compress and forward features to a
central server without access to the other nodes' inputs. This
\emph{distributed multiview representation learning} setting is illustrated
in Fig.~\ref{fig:distributed} and formalized in Section~\ref{sec:setup}.
Unlike federated learning~\cite{mcmahan2017}, where clients hold
independent datasets, every node here observes a different \emph{view} of
the same underlying instance, and all views are jointly required at
inference time.

The prohibition on inter-encoder communication introduces a fundamental
trade-off. Each encoder must compress its view aggressively to keep the joint
representation complexity under control and limit the generalization error,
yet the compressed features must collectively be sufficient for the decoder
to reconstruct the label, without any encoder knowing what the others will
contribute. This trade-off raises two intertwined questions: (i) from a
generalization standpoint, is it preferable for encoders to produce
\textit{redundant} representations that overlap across views, or
\textit{complementary} ones that partition the information? and (ii) what regularization objective steers encoders toward the statistically superior
regime during training?

Existing work on distributed representation learning has addressed these
questions almost exclusively by extending the IB
framework~\cite{estella2018distributed,wang2019deep,federici2020learning,aguerri21,wang2021cross,moldoveanu2021network,wan2021multi,lin2022dual,huang2022multi,cui2024novel,yan2024differentiable,huang2024generalized};
see~\cite{goldfeld2018estimating,zaidi2020information,yan2021deep,hu2024survey}
for surveys. Because all of these methods inherit the assumption that mutual
information is a meaningful proxy for generalization, an assumption
contradicted by both theory and
experiment~\cite{kolchinsky2018caveats,rodriguez2019information,amjad2019learning,geiger2019information,dubois2020learning,lyu2023recognizable,sefidgaran2023minimum},
they offer no rigorous generalization guarantees.

The present paper takes a different route. We study the distributed
multiview problem directly through the lens of generalization error,
deriving bounds that simultaneously illuminate what good representations
look like and motivate a principled new family of regularizers.

\textbf{Contributions.} Our main contributions are as follows.

\begin{itemize}[leftmargin=0.9em,topsep=0pt]
  \setlength\itemsep{0.1 cm}

\item \textbf{Generalization bounds.} We derive several bounds on the
generalization error of the distributed multiview setting of
Fig.~\ref{fig:distributed}. All bounds are expressed in terms of
$\textnormal{MDL}(\vc{Q})$, the Minimum Description Length of the joint
latent variables relative to a data-dependent symmetric prior $\vc{Q}$.
As argued in~\cite{sefidgaran2023minimum,futami2025information}, in sharp contrast with MI-based complexity measures $\textnormal{MDL}(\vc{Q})$ reflects the structure and captures the `simplicity' of the encoder.
\begin{itemize}
    \item For \textbf{classification}, we adapt the single-encoder bound
of~\cite{sefidgaran2023minimum} to the $K$-encoder setting via a suitable
reformulation of the argument, yielding a bound of order
$\sqrt{\textnormal{MDL}(\vc{Q})/n}$.
    \item For \textbf{regression}, we establish a new bound of the same
    order. When specialized to $K=1$, $Y=X$, and discrete latent variables
    (VQ-VAE), it recovers the bound of~\cite{futami2025information} with a
    sharper constant; more importantly, the proof makes no use of the
    discrete lattice structure and handles continuous latent spaces, requiring several new technical ingredients.
    The main one is a concentration bound for a \emph{bilinear permutation
    statistic} pairing the centered targets with the centered decoder
    outputs. Such a statistic has dependent terms: a uniform
    permutation samples without replacement. Hence, the sign-symmetrization
    argument that is sufficient in the classification case breaks down. Also, the
    codebook-partitioning trick of~\cite{futami2025information} is not applicable for the continuous latent space cases. We obtain the required
    sub-Gaussian estimate through Stein's method for exchangeable
    pairs~\cite{chatterjee2006stein}; see Lemma~\ref{lem:stein_bound}.
    \item We further develop a \textbf{structure-aware bound} that
    explicitly reflects the distributed architecture, by showing that
    $\textnormal{MDL}(\vc{Q})$ is upper-bounded by a quantity
    $\textnormal{MDL}_{dist}(\vc{Q}_1,\ldots,\vc{Q}_K)$, defined
    in~\eqref{eq:Dist_Redundancies2}, which decomposes into per-view
    marginal MDL terms minus a non-negative joint term. The negative sign
    of this joint term implies that cross-view statistical correlations in
    the extracted representations can only tighten the bound, providing the
    first generalization-theoretic justification for the empirically
    observed benefits of cross-view feature alignment.
\end{itemize}

\item \textbf{Marginal-only regularizers.} Combining the above bounds with
the Gaussian mixture prior of~\cite{sefidgaran2025generalization}, we show
that per-view regularizers can be deployed independently at each encoder,
for both classification and regression tasks. While this straightforwardly
improves over unregularized training, it incurs a structural cost: each
cross-view redundancy is penalized once per participating encoder, actively
discouraging the overlapping representations that the structure-aware bound
identifies as beneficial.

\item \textbf{Gaussian product mixture regularizer.} To remove this
tension, we construct a joint prior over all views as a mixture of Gaussian
products. This prior satisfies three key properties simultaneously:
\textbf{i.}~its per-view marginal is a Gaussian mixture, remaining locally
consistent with~\cite{sefidgaran2025generalization}; \textbf{ii.}~it
assigns a lower penalty to cross-view redundancies than any product of
marginal priors, in direct alignment with the structure-aware bound;
\textbf{iii.}~its parameters can be updated in a fully distributed fashion
with minimal server-client communication overhead. In the lossy variant, the
resulting update equations take the form of a weighted distributed attention
mechanism.

\item \textbf{Experiments.} We validate the theoretical findings through
extensive experiments covering both classification and regression tasks,
multiple encoder architectures (CNN4, ResNet), between 2 and 8 views, and
several benchmark datasets. Our method, GPM-MDL, consistently and
substantially outperforms both unregularized training and the per-view VIB
baseline across all tested configurations.
\end{itemize}

\textbf{Notations.} Random variables and their realizations are denoted by
upper and lower case letters, respectively, and their support sets by
calligraphic fonts, \eg $X$, $x$, and $\mathcal{X}$. The distribution of
$X$ is denoted by $P_X$, assumed to be a probability mass function for
discrete support and a probability density function otherwise. The
Kullback-Leibler divergence between distributions $P$ and $Q$ is
$D_{KL}(P\|Q)\coloneqq\mathbb{E}_P[\log(P/Q)]$ if $P\ll Q$, and $\infty$
otherwise. We denote $\{1,\ldots,n\}$ by $[n]$ for $n\in\mathbb{N}$.

\section{Problem setup} \label{sec:setup}
We consider a distributed $C$-class $K$-view classification and regression setups, as described below.

\textbf{Data.} We assume that the \emph{input data} $Z$, which takes value in the \emph{input space} $\mathcal{Z}$ according to an unknown distribution $\mu$, is composed of two parts $Z=(X,Y)$, where \textbf{i.} $X$ represents the vector of \emph{features} of the input data, taking values in the \emph{feature space} $\mathcal{X}$. More precisely, $X=(X_1,\ldots,X_K)$, where $X_k \in \mathcal{X}_k$, $k\in[K]$ corresponds to the $k$'th view of the data. Note that $\mathcal{X} = \mathcal{X}_1 \times \cdots \times \mathcal{X}_K$. We assume that each view $X_k$ of the features is distributed according to $\mu_{X_k}$, and the full vector of features $X$ is distributed according to $\mu_{X^K}\coloneqq \mu_{X}$. \textbf{ii.} $Y\in \mathcal{Y}$ represents, in the classification task, a label ranging from 1 to $C$ (\ie $\mathcal{Y}=[C]$), or, in the regression task, a target which is a bounded $d_y$-dimensional vector as stated in the assumption below. 

\begin{assumption} \label{assum:bounded} There exists $R\in \mathbb{R}_+$ such that $\forall y,y' \in \mathcal{Y}$, $\|y-y'\|^2 \leq R$.  
\end{assumption}

We denote the underlying distribution of $Y$ by $\mu_Y$. We further denote the joint distribution of the features and the label by $\mu \coloneqq \mu_{X|Y} \mu_Y \coloneqq \mu_{X} \mu_{Y|X}$. 

\textbf{Training dataset.} To learn a model, we assume the availability of a \emph{training dataset} $S=\{Z_1,\ldots,Z_n\}\sim \mu^{\otimes n}\eqqcolon P_S$, composed of $n$ i.i.d. samples $Z_i=(X_i,Y_i)$ of the input data. Note that each $X_i$, $i\in[n]$, is composed of $K$ views, \ie $X_i = (X_{i,1},\ldots,X_{i,K})$. In our analysis, we often use a \emph{test} or \emph{test} dataset $S'=\{Z'_1,\ldots, Z'_n\}\sim \mu^{\otimes n}\eqqcolon P_{S'}$, where $Z'_i=(X'_i,Y'_i)$ and $X'_i = (X'_{i,1},\ldots,X'_{i,K})$. Furthermore, we denote the restrictions of sets $S$ and $S'$ to view $k$ by $S_k = \big\{(X_{1,k},Y_1),\ldots,(X_{n,k},Y_n)\big\}$ and $S'_k = \big\{(X'_{1,k},Y'_1),\ldots,(X'_{n,k},Y'_n)\big\}$, respectively. To simplify the notation, we denote the features and labels of $S$ and $S'$ by $\vc{X}\coloneqq X^n\sim \mu_X^{\otimes n}$, $\vc{Y}\coloneqq Y^n \sim \mu_Y^{\otimes n}$, $\vc{X}'\coloneqq X^{\prime n} \sim \mu_X^{\otimes n}$ and $\vc{Y}'\coloneqq Y^{\prime n} \sim \mu_Y^{\otimes n}$, respectively. Similar notations are used to denote the $k$'th view of $S$ and $S'$ by $\vc{X}_k\coloneqq X^n_k$ and $\vc{X}'_k\coloneqq X^{\prime n}_k$, respectively.

\textbf{Distributed setup.} We assume that there are $K$ clients, each observing a single view. Client $k\in[K]$, upon observing view $X_k$ and having access to encoder $w_{e,k}\in \mathcal{W}_{e,k}$, produces the \emph{representation} or \emph{latent variable} $U_k \in \mathcal{U}_k$, possibly stochastically. We denote the collection of all latent variables as $U = (U_1,\ldots,U_K) \in \mathcal{U} = \mathcal{U}_1 \times \cdots \times \mathcal{U}_K$, where for simplicity $\mathcal{U}_1 = \cdots =\mathcal{U}_K = \mathbb{R}^{d}$ for some $d\in \mathbb{N}^*$. Similarly, we denote the collection of all encoders by $\vc{w}_e = (w_{e,1},\ldots,w_{e,K})\in \mathcal{W}_e= \mathcal{W}_{e,1} \times \cdots \times \mathcal{W}_{e,K}$. These latent variables are forwarded to a central server, which, using decoder $w_d\in \mathcal{W}_{d}$, produces the prediction $\hat{Y}$ of $Y$. We write $w\coloneqq (w_e,w_d)\in \mathcal{W} = \mathcal{W}_e \times \mathcal{W}_d$ for the full model. The setup is illustrated in Fig.~\ref{fig:distributed}.

\textbf{Learning algorithm.} We consider a general stochastic learning framework in which the learning algorithm $\mathcal{A}\colon \mathcal{Z}^n \to \mathcal{W}$, upon observing training dataset $S$, outputs a model $\mathcal{A}(S)=W\in \mathcal{W}$ consisting of $K$ encoders $(W_{e,1},\ldots,W_{e,K}) \eqqcolon W_e$ and a decoder $W_d$. The distribution induced by $\mathcal{A}$ is denoted by $P_{W|S}=P_{W_e,W_d|S}$. The joint distribution of $(S,W)$ is denoted by $P_{S,W}$, and the marginal distribution of $W$ is denoted by $P_W$. Furthermore, we denote the induced conditional distribution of the latent variables $U$ given the encoder and input by $P_{U|X,W_e} = \bigotimes \nolimits_{k\in[K]} P_{U_k|X_k,W_{e,k}}$. The conditional distribution of the prediction $\hat{Y}\in \hat{\mathcal{Y}} = \mathcal{Y}$ given the decoder and latent variables is denoted by $P_{\hat{Y}|U,W_d}$, and satisfies $P_{\hat{Y}|X,W} = \E_{U\sim P_{U|X,W_e}}\big[ P_{\hat{Y}|U,W_d}\big]$. Throughout, we let $\vc{U}\coloneqq U^n$ and $\vc{U}'\coloneqq U^{\prime n}$ and we use the shorthand notation 
\begin{align*}
    P_{\vc{U}, \vc{U}'|\vc{X},\vc{X}',W_e} \coloneq \bigotimes\nolimits_{i\in[n]} \left\{P_{U_i|X_i,W_e} P_{U'_i|X'_i,W_e}\right\}.
\end{align*}
\textbf{Classification loss.} The quality of a model $w$ for the classification task is measured by the 0-1 loss $\ell_b\colon \mathcal{Z} \times \mathcal{W} \to [0,1]$:
\begin{align*}
    \ell_b(z,w) \coloneqq \E_{\hat{Y}\sim P_{\hat{Y}|x,w}} \left[\mathbbm{1}\{y \neq \hat{Y}\}\right] = \E_{U\sim P_{U|x,w_e}}\E_{\hat{Y}\sim P_{\hat{Y}|U,w_d}} \left[\mathbbm{1}\{y \neq \hat{Y}\}\right].
\end{align*}
\textbf{Regression loss.} For the regression task, we consider the squared error loss $\ell_r\colon \mathcal{Z} \times \mathcal{W} \to \mathbb{R}_+$:
\begin{align*}
    \ell_r(z,w) \coloneqq \E_{\hat{Y}\sim P_{\hat{Y}|x,w}} \left[\|y - \hat{Y}\|^2\right] = \E_{U\sim P_{U|x,w_e}}\E_{\hat{Y}\sim P_{\hat{Y}|U,w_d}}  \left[\|y - \hat{Y}\|^2\right].
\end{align*}
In what follows, we use the general notation $\ell(z,w)$, which is specialized to $\ell_b(z,w)$ and $\ell_r(z,w)$ for the classification and regression tasks, respectively.

\textbf{Risks.} The goal is to find a model minimizing the \emph{population risk} $\mathcal{L}(w)=\E_{Z\sim \mu}\left[\ell(Z,w)\right]$. Since $\mu$ is unknown, one minimizes instead the \emph{empirical risk} $\hat{\mathcal{L}}(s,w)=\frac{1}{n}\sum_{i\in[n]}\ell(z_i,w)$. The gap between these two quantities is the \emph{generalization error}:
\begin{align}
    \gen(s,w) \coloneqq \mathcal{L}(w) - \hat{\mathcal{L}}(s,w).
\end{align}
When needed, we use the specialized notations $\big(\mathcal{L}_b,\hat{\mathcal{L}}_b,\gen_b\big)$ and $\big(\mathcal{L}_r,\hat{\mathcal{L}}_r,\gen_r\big)$ for the classification and regression tasks, respectively.
\begin{definition}[Conditional symmetric prior {\cite{sefidgaran2023minimum}}] \label{def:symmetry} 
A conditional prior $Q(U^{2n}|Z^{2n},W_e)$ is said to be symmetric if $Q(U^{2n}_{\pi}|Z^{2n},W_e)$ is invariant under all permutations $\pi\colon [2n]\mapsto [2n]$ satisfying $\forall i \colon Y^{2n}_{i}{=}Y^{2n}_{\pi(i)}$.
\end{definition}

\begin{definition}[Symmetric prior {\cite{futami2025information}}] \label{def:symmetry_reg} 
A conditional prior $Q(U^{2n}|Z^{2n},W_e)$ is said to be symmetric if $Q(U^{2n}_{\pi}|Z^{2n},W_e)$ is invariant under all permutations $\pi\colon [2n]\mapsto [2n]$.
\end{definition}

\textbf{Encoders with Gaussian-distributed latent variables.} For the practical parts of this paper, we assume that the encoder for an input $x$ outputs a mean vector $\mu_x=(\mu_{x,1},\ldots,\mu_{x,K})\in \mathbb{R}^{Kd}$ and a standard deviation vector $\sigma_x=(\sigma_{x,1},\ldots,\sigma_{x,K})\in \mathbb{R}^{Kd}$. For every $k\in[K]$, the latent variable $U_k$ is assumed to follow a multivariate Gaussian distribution with diagonal covariance, \ie $U_k \sim \mathcal{N}\big(\mu_{x,k},\diag(\sigma_{x,k}^2)\big)$, where $\diag(\sigma_{x,k}^2)$ denotes a $d\times d$ diagonal matrix with diagonal entries equal to $\sigma_{x,k}^2$. Given the pair $(x,w_e)$, the latent variables $(U_1,\ldots,U_K)$ are conditionally independent of each other. Under these assumptions,
\begin{equation}
    P_{\vc{U}, \vc{U}'|\vc{X},\vc{X}',W_e} = \bigotimes\nolimits_{i\in[n]}\bigotimes\nolimits_{k\in[K]} \Big\{\mathcal{N}\big(\mu_{x_i,k},\diag(\sigma_{x_i,k}^2)\big)\mathcal{N}\big(\mu_{x'_i,k},\diag(\sigma_{x'_i,k}^2)\big)\Big\}. \label{def:gaussian_encoders}
\end{equation}

\section{Structure-agnostic generalization bounds and regularizers for multiview representation learning} \label{sec:Bounds_multi_view}
We establish generalization bounds for both classification and regression in terms of the \emph{Minimum Description Length} $\textnormal{MDL}(\mathbf{Q})$ of the latent variables. Intuitively, $\textnormal{MDL}(\mathbf{Q})$ quantifies how many bits are needed on average to encode the latent variables under prior $\mathbf{Q}$: encoders with simpler, more structured outputs require fewer bits and thus generalize better.

A key advantage over mutual information- and PAC-Bayes-based complexity measures \cite{xu2017information,steinke2020reasoning,mcallester1998some,catoni2003pac,catoni2007pac} is that $\textnormal{MDL}(\mathbf{Q})$ is data- and model-dependent, which makes it directly usable as a training-time regularizer: at each iteration a mini-batch gives a tractable estimate of latent complexity. In contrast, model-complexity bounds depend on the model parameters, of which only a single training trajectory is available, making it impossible to obtain statistically reliable estimates of the required complexity terms during optimization.

We begin by bounding the generalization error of the multiview system as a whole, treating it as a single centralized algorithm and ignoring the distributed structure among the encoders. Combining these bounds with the Gaussian mixture prior construction of \cite{sefidgaran2025generalization} then yields a simple per-view regularizer, whose limitations motivate the structure-aware joint prior developed in Section~\ref{sec:joint_multivew_regularizer}.

\subsection{Classification}
A generalization upper bound for representation learning algorithms in terms of the MDL of the latent variables was established in \cite[Theorem~1]{sefidgaran2023minimum} for the single-view case. This bound extends directly to the multiview case by considering the joint MDL of all latent variables across all views.
\begin{theorem}[{\cite[Theorem~4]{sefidgaran2023minimum}}]\label{th:generalizationExp_old}
Consider a $C$-class $K$-view classification problem and a learning algorithm $\mathcal{A}\colon \mathcal{Z}^n\to \mathcal{W}$ that induces the joint distribution $(S,S',\vc{U},\vc{U}',W) \sim P_{S'} P_{S,W} P_{\vc{U}|\vc{X},W_e}P_{\vc{U}'|\vc{X}',W_e}$. Then, for any conditionally symmetric prior $\vc{Q}(\vc{U},\vc{U}'|S,S',W_e)$,
\begin{align}
\E_{\vc{S},W}\left[\gen_b(S,W)\right] \leq \sqrt{\frac{2\,\textnormal{MDL}(\vc{Q})+C+2}{n}}, \label{eq:exp_bound_classification}
\end{align}
where
\begin{align}
  \textnormal{MDL}(\vc{Q}) \coloneqq \E_{S,S',W_e} \big[ D_{KL}\big(P_{\vc{U}, \vc{U}'|\vc{X},\vc{X}',W_e} \big\| \vc{Q}\big) \big]. \label{eq:MDL_original}
\end{align}
\end{theorem}
This result establishes a bound of order $\sqrt{\textnormal{MDL}(\vc{Q})/n}$. As shown in \cite[Theorem~1]{sefidgaran2025generalization}, this dependence can be tightened to order $\textnormal{MDL}(\vc{Q})/n$. Notably, the bound depends only on the encoder, which is consistent with the encoder-decoder design principle: the encoder extracts representations that control generalization, while the decoder minimizes empirical risk. Moreover, the prior is both data- and model-dependent, which, as discussed in the next sections, allows the inherent structure of the representations to be discovered and reinforced simultaneously.

\subsection{Regression}
A similar result holds for the regression task.
\begin{theorem}\label{th:generalizationExp_reg}
Suppose Assumption~\ref{assum:bounded} holds for some $R\in \mathbb{R}_+$. Consider a $K$-view regression problem and a learning algorithm $\mathcal{A}\colon \mathcal{Z}^n\to \mathcal{W}$ inducing $(S,S',\vc{U},\vc{U}',W) \sim P_{S'} P_{S,W} P_{\vc{U}|\vc{X},W_e}P_{\vc{U}'|\vc{X}',W_e}$. Then, for any symmetric prior $\vc{Q}(\vc{U},\vc{U}'|S,S',W_e)$,
\begin{align}
\E_{\vc{S},W}\left[\gen_r(S,W)\right] \leq R\sqrt{\frac{8\,\textnormal{MDL}(\vc{Q})}{n}}+\frac{R\sqrt{2}}{\sqrt{n}}, \label{eq:exp_bound_regression}
\end{align}
where $\textnormal{MDL}(\vc{Q})$ is as defined in \eqref{eq:MDL_original}.
\end{theorem}

The bound of \cite{futami2025information} covers the special case $K=1$, $Y=X$, with discrete-valued latent variables in a VQ-VAE setup. Theorem~\ref{th:generalizationExp_reg} slightly improves the constant from $9$ to $8$ for this special case and more importantly, extends this to arbitrary $K$, arbitrary joint distributions of $(X,Y)$, and \emph{continuous latent variables}, with a proof technique that does not rely on the discreteness of the latent space. This is in contrast to the proof of \cite{futami2025information}, which relies strongly on the discreteness of the latent space. We use multiple new ideas to handle the general case, including Stein's method for exchangeable variables \cite{chatterjee2006stein}. We first discuss the statement, and then sketch the proof, emphasizing the two symmetrization steps on which it rests.

The right-hand side of \eqref{eq:exp_bound_regression} splits into a \emph{model-dependent} and a \emph{model-independent} part. The first term of \eqref{eq:exp_bound_regression}, $R\sqrt{8\,\textnormal{MDL}(\vc{Q})/n}$, is governed \emph{solely by the encoders}: as in Theorem~\ref{th:generalizationExp_old}, the decoder $W_d$ does not appear, so the encoder-decoder division where the encoder controls complexity, the decoder fits the empirical risk, carries over similarly for the regression case. The second term, $R\sqrt{2/n}$, is independent of the algorithm, of $K$, and of the latent dimension; it originates from the $\Theta(n^{-1/2})$ fluctuation of the empirical target mean $\frac{1}{n}\sum_i Y_i$ around $\E_Y[Y]$ and therefore \emph{cannot} be reduced by compressing the representations. In particular, in the regime $\textnormal{MDL}(\vc{Q})\to 0$ the bound saturates at $R\sqrt{2/n}$ rather than vanishing, in analogy with the $\sqrt{(C+2)/n}$ floor of Theorem~\ref{th:generalizationExp_old}. Finally, since Assumption~\ref{assum:bounded} bounds the \emph{squared} target diameter by $R$, both terms scale linearly in the scale of the squared loss, as they should.

\textbf{Sketch of proof of Theorem~\ref{th:generalizationExp_reg}.}
The full argument is given in Section~\ref{pr:generalizationExp_reg}; we outline it here in four steps.

\emph{Step 1 (Test-sample decomposition).} Introducing the test dataset $S'$ and the auxiliary losses
$\ell_a(y',\hat{y}',\hat{y})=\|\hat{y}'-y'\|^2-\|\hat{y}-y'\|^2$ and $\ell_b(\hat{y},y',y)=\|\hat{y}-y'\|^2-\|\hat{y}-y\|^2$,
the expected generalization error splits as $\E_{S,W}[\gen_r(S,W)]=\Delta_1+\Delta_2$, where

\begin{itemize}
    \item $\Delta_1$ scores the test prediction $\hat{Y}'_i$ and the training prediction $\hat{Y}_i$ against the \emph{same} target $Y'_i$, and therefore measures a train/test asymmetry carried by the latent variables alone; it is handled by exchanging $U_i$ and $U'_i$ and due to the fact that $\ell_a$ is antisymmetric in its two latent arguments.
    \item The term $\Delta_2$ scores the \emph{same} prediction $\hat{Y}_i$ against its own target $Y_i$ and against an independent copy $Y'_i$, and therefore measures the alignment between a target and the prediction formed at the \emph{same} sample index. This term is handled by breaking that index matching, \ie by permuting the latent indices while holding the targets in place.
\end{itemize}
   Thus, in both cases the targets are held fixed; what differs is the axis along which the randomization acts: the train/test axis for $\Delta_1$, the sample axis for $\Delta_2$. This decomposition is inspired by \cite{futami2025information}; both terms are then analyzed differently, since the arguments of \cite{futami2025information} for either term use the ``VQ'' step, \ie the discreteness of the latent alphabet.

\emph{Step 2 ($\Delta_1$).} A change of measure from $P_{\vc{U},\vc{U}'|\vc{X},\vc{X}',W_e}$ to $\vc{Q}$ (Donsker-Varadhan) pays $\textnormal{MDL}(\vc{Q})$ and leaves the cumulant generating function of $\frac{1}{n}\sum_i \ell_a(U'_i,U_i|Y'_i)$ under $\vc{Q}$. The key observation is that $\ell_a(\cdot,\cdot|Y'_i)$ is \emph{antisymmetric} in its two latent arguments and bounded by $R$ in absolute value. Since $\vc{Q}$ is symmetric, the pair $(U_i,U'_i)$ may be swapped independently across $i\in[n]$ according to i.i.d.\ Rademacher signs without changing the expectation. This produces a sum of \emph{independent}, bounded, symmetric terms, and Hoeffding's lemma yields the sub-Gaussian proxy $R^2/n$.

\emph{Step 3 ($\Delta_2$).} Expanding the squares and using that $\vc{Y}$ and $\vc{Y}'$ are identically distributed reduces $\Delta_2$ to $\frac{2}{n}\sum_i\langle Y_i-\E_Y[Y],g(U_i,W_d)\rangle$, where $g(u,w_d)\coloneqq\E[\hat{Y}\,|\,u,w_d]$ is the posterior mean produced by the decoder. Writing $g(U_i,W_d)=\bar{g}(\vc{U},W_d)+C_i$ with $\bar{g}$ the empirical mean of the $g(U_i,W_d)$'s, the $\bar{g}$-part is a purely statistical quantity, bounded by $2\sqrt{R}\,\E\|\frac{1}{n}\sum_i(Y_i-\E_Y[Y])\|\leq R\sqrt{2/n}$ via Cauchy-Schwarz and Jensen; this is the second term of the bound. Centering the targets as $V_i\coloneqq Y_i-\frac{1}{n}\sum_j Y_j$, what remains is the bilinear pairing
\begin{align}
    \frac{1}{n}\sum_{i=1}^n \langle V_i, C_i\rangle, \qquad \textstyle\sum_i V_i = \sum_i C_i = 0, \label{eq:bilinear_pairing}
\end{align}
between two \emph{centered} sequences, one carried by the targets and one by the latent variables.

\emph{Step 4 ($\Delta_2$: Stein's method for exchangeable pairs).} After a second change of measure, one needs the cumulant generating function of \eqref{eq:bilinear_pairing} under $\vc{Q}$. Exchangeability of $\vc{Q}$ in the training indices allows replacing $C_i$ by $C_{\pi(i)}$ for a uniformly random permutation $\pi$ of $[n]$, independent of everything else, at no cost. Unlike Step~2, however, this randomization does \emph{not} produce independent terms: a uniform permutation samples \emph{without} replacement, so $\{\langle V_i,C_{\pi(i)}\rangle\}_{i\in[n]}$ are dependent and neither Hoeffding's lemma nor the bounded-difference inequality applies. This is precisely the point at which \cite{futami2025information} exploits the discreteness of the VQ codebook to group the indices assigned to a common codeword. However, this trick cannot be applied for continuous latent spaces.

We resolve this by Chatterjee's method of exchangeable pairs \cite{chatterjee2006stein}. Set $\mathsf{X}\coloneqq\sum_{i}\langle V_i,C_{\pi(i)}\rangle$ and let $\mathsf{X}'$ be obtained from $\mathsf{X}$ by transposing the images of two independent uniform indices $I,J$; then $(\mathsf{X},\mathsf{X}')$ is an exchangeable pair. Taking $f(\mathsf{x})=\mathsf{x}$ and the antisymmetric kernel $F(\mathsf{x},\mathsf{y})=\frac{n}{2}(\mathsf{x}-\mathsf{y})$, a direct computation shows the Stein identity $\E[F(\mathsf{X},\mathsf{X}')\,|\,\mathsf{X}]=f(\mathsf{X})$. Chatterjee's result then controls the moment generating function of $\mathsf{X}$ through the single quantity
\begin{align}
    \Delta(\mathsf{X}) = \frac{1}{4n}\sum_{i=1}^n\sum_{j=1}^n\big\langle V_i-V_j,\,C_{\pi(i)}-C_{\pi(j)}\big\rangle^2 \leq \frac{nR^2}{4},
\end{align}
where the inequality follows from Cauchy-Schwarz together with $\|V_i-V_j\|^2=\|Y_i-Y_j\|^2\leq R$ and $\|C_i-C_j\|^2=\|g(U_i,W_d)-g(U_j,W_d)\|^2\leq R$, both consequences of Assumption~\ref{assum:bounded}. Since this bound is deterministic, $\|\Delta(\mathsf{X})\|_{\infty}\leq nR^2/4$, and one recovers the \emph{same} sub-Gaussian proxy $R^2/n$ as in Step~2. Balancing the two changes of measure with the common parameter $\lambda=\sqrt{2n\,\textnormal{MDL}(\vc{Q})}/R$ completes the proof. \hfill$\square$

\subsection{Lossy generalization bounds} \label{sec:lossy}

The bounds above assume lossless encoding but admit a natural extension to a \textit{lossy} regime, which is used in our regularizer. Beyond greater generality, lossy bounds offer two key advantages: they remain non-vacuous for deterministic encoders and account for the empirically observed \emph{geometrical compression} phenomenon \cite{geiger2021information}. By contrast, MI-based bounds \cite{xu2017information} suffer from both shortcomings \cite{haghifam2023limitations,livni2023information}, which the lossy approach circumvents \cite{sefidgaran2024data,sefidgaran2025generalization}. For brevity, only the lossy extension of Theorem~\ref{th:generalizationExp_old} is stated; a similar extension holds for Theorem~\ref{th:generalizationExp_reg}.

Let $\hat{W}_e\in \mathcal{W}_e$ be a quantized encoder defined by an arbitrary conditional distribution $P_{\hat{W}_e|S}$ satisfying the \emph{distortion} criterion $\E_{P_{S,W}P_{\hat{W}_e|S}}\left[\gen_b(S,W)-\gen_b(S,\hat{W})\right] \leq \epsilon$, where $\hat{W}=(\hat{W}_e,W_d)$. Then,
\begin{equation}
    \E_{\vc{S},W}\left[\gen_b(S,W)\right] \leq \sqrt{\frac{2\,\textnormal{MDL}(\vc{Q})+C+2}{n}}+\epsilon, \label{eq:lossy_bound}
\end{equation}
where $\textnormal{MDL}(\vc{Q})$ is evaluated at the quantized encoder:
\begin{equation}
  \textnormal{MDL}(\vc{Q}) \coloneqq \E_{S,S',\hat{W}_e} \Big[ D_{KL}\Big(P_{\vc{U}, \vc{U}'|\vc{X},\vc{X}',\hat{W}_e} \big\| \vc{Q}(\vc{U},\vc{U}'|S,S',\hat{W}_e)\Big) \Big]. \label{eq:MDL_original_quantized}
\end{equation}
Introducing lossy compression ($\epsilon > 0$) can reduce $\textnormal{MDL}(\vc{Q})$ inside the square root in~\eqref{eq:lossy_bound}, at the cost of the additive residual $\epsilon$. When the reduction outweighs this cost, the lossy bound~\eqref{eq:lossy_bound} is tighter than its lossless counterpart in Theorem~\ref{th:generalizationExp_old}. We refer the reader to \cite{sefidgaran2024data} for a concrete example in a related setup, and to \cite[Appendix~A]{sefidgaran2025generalization} for the proof and further discussion.

\subsection{Marginal-only regularizers} \label{sec:marginal_reg}

The discussion here focuses on the classification task; an analogous treatment applies to regression. For $k\in[K]$, let $\vc{Q}_k(\vc{U}_k,\vc{U}'_k|S_k,S'_k,W_{e,k})$ be per-view conditionally symmetric priors, and define
\begin{align}
    \vc{Q}(\vc{U},\vc{U}'|S,S',W_e) \coloneqq \bigotimes\nolimits_{k\in[K]} \vc{Q}_k(\vc{U}_k,\vc{U}'_k|S_k,S'_k,W_{e,k}).
\end{align}
It is straightforward to verify that $\vc{Q}$ is also a conditionally symmetric prior, and that
\begin{align}
     \textnormal{MDL}(\vc{Q}) = \sum_{k\in[K]} \textnormal{MDL}(\vc{Q}_k),
\end{align}
where $\textnormal{MDL}(\vc{Q}_k) \coloneqq \E_{S_k,S'_k,W_{e,k}} \big[ D_{KL}\big(P_{\vc{U}_k, \vc{U}'_k|\vc{X}_k,\vc{X}'_k,W_{e,k}} \big\| \vc{Q}_k\big) \big]$.

This decomposition allows each client $k$ to use $\textnormal{MDL}(\vc{Q}_k)$ as a local regularizer, depending only on its own encoder. In particular, each client can independently apply the data-dependent Gaussian mixture regularizer of \cite{sefidgaran2025generalization}, summarized in Appendix~\ref{sec:single_view} for completeness. However, as discussed in Section~\ref{sec:Bounds_dist_multi_view}, this marginal approach has a notable drawback: redundant features shared across views are penalized multiple times, implicitly discouraging the encoders from extracting cross-view redundancies.

\section{Structure-aware generalization bounds for multiview representation learning} \label{sec:Bounds_dist_multi_view} 

The setup-agnostic bounds of the previous section treat the multiview system as a single centralized algorithm and do not reflect its distributed architecture. To address this, we derive an upper bound on $\textnormal{MDL}(\vc{Q})$ that explicitly accounts for the distributed structure. The result is stated for the classification task; an analogous result holds for regression.%

Denote $\tilde{Y}^{2n}\coloneqq (\vc{Y},\vc{Y}')$ and for a given $\tilde{Y}^{2n}$, let $Y_i =\tilde{Y}_i$ and $Y'_i=\tilde{Y}_{i+n}$, with similar notations for $\tilde{U}^{2n}$. For a given $\tilde{Y}^{2n}$, let the permutation $\pi_{\tilde{Y}^{2n}}\colon[2n]\to [2n]$, denoted simply as $\pi$, satisfy: \textbf{i.} for $i\in[n]$, $\pi(i)\in \{i\} \cup \{n+1,\ldots,2n\}$ and $\pi(i+n)\in \{1,\ldots,n\} \cup \{i+n\}$; \textbf{ii.} $\pi(\pi(i))=i$; \textbf{iii.} $\tilde{Y}_i = \tilde{Y}_{\pi(i)}$; and \textbf{iv.} it maximizes the cardinality of $\{i\colon \pi(i) \neq i\}$. If multiple such permutations exist, one is chosen deterministically. Define
\begin{align}
\overline{\mathsf{P}}_{\vc{U},\vc{U}'|\vc{Y},\vc{Y}',W_e} \coloneqq \E_{\vc{X},\vc{X}'|\vc{Y},\vc{Y}',W_e}\left[\big(P_{\tilde{U}^{2n}|\vc{X},\vc{X}',W_e}+P_{\tilde{U}^{2n}_{\pi}|\vc{X},\vc{X}',W_e}\big)\big/2\right]. \label{eq:sym_prior_dist}
\end{align}
The following result is proved in Section~\ref{pr:distributed_gen}.

\begin{theorem} \label{th:distributed_gen}
Consider the setup of Theorem~\ref{th:generalizationExp_old}. Let, for every $k \in [K]$, $\vc{Q}_k$ be some conditionally symmetric prior for view $\mathbf{X}_k$. Then,
\begin{equation}
\E_{\vc{S},W}\left[\gen_b(S,W)\right] \leq \sqrt{\frac{2\,\textnormal{MDL}_{dist}(\vc{Q}_1,\ldots,\vc{Q}_K)+C+2}{n}},
\end{equation}
where
\begin{align}
\textnormal{MDL}_{dist}(\vc{Q}_1,\ldots,\vc{Q}_K) \coloneqq &\sum\nolimits_{k\in[K]} \E_{S_k,S'_k,W_{e,k}} \left[ D_{KL}\left(P_{\vc{U}_k, \vc{U}'_k|\vc{X}_k,\vc{X}'_k,W_{e,k}} \Big\| \vc{Q}_k\right)\right] \nonumber\\
&- \E_{S,S',W_e}\left[D_{KL} \Big(\overline{\mathsf{P}}_{\vc{U},\vc{U}'|\vc{Y},\vc{Y}',W_e}\big\|\prod\nolimits_{k\in[K]}\vc{Q}_k\Big)\right].\label{eq:Dist_Redundancies2}
\end{align}
\end{theorem}

The two terms in $\textnormal{MDL}_{dist}$ play complementary roles: the first captures the marginal MDL of each view, while the second, which enters with a negative sign, captures the joint MDL of all views jointly. Crucially, the second term increases when the extracted representations $(\mathbf{U}_1,\ldots,\mathbf{U}_K)$ are statistically correlated, implying that statistical correlations help to tighten the bound. This provides a partial but rigorous answer to the coordination question posed in the introduction: encoders benefit, in terms of generalization error, from learning representations that are statistically redundant across views. To the best of our knowledge, this is the first result to address this question from a generalization error perspective. The related work~\cite{aguerri21} studied a distributed version of the Information Bottleneck method, with an arbitrary number of encoders. Among other results, the authors use a connection with a multi-terminal source coding problem under logarithmic loss measure to establish a distributed version of 
Tishby's relevance-complexity region~\cite{tishby2000information}. For a comparison with~\eqref{eq:Dist_Redundancies2}, for simplicity, we here restrict to $K=2$ views. In this case, the (sum) information complexity term, used as a regularizer therein, is
\begin{equation}
R_1 + R_2 = I(\mathbf{U}_1;\mathbf{X}_1|\mathbf{Y}) + I(\mathbf{U}_2;\mathbf{X}_2|\mathbf{Y}) + I(\mathbf{U}_1, \mathbf{U}_2; \mathbf{Y}).
\label{eq:sum-rate-DIB}
\end{equation}
 Through straightforward algebra, it can be shown that with the Markov Chain $\mathbf{U}_1 \leftrightarrow \mathbf{X}_1 \leftrightarrow \mathbf{Y} \leftrightarrow \mathbf{X}_2 \leftrightarrow \mathbf{U}_2$ that is assumed therein the RHS of \eqref{eq:sum-rate-DIB} can be written equivalently as
 \begin{equation}
 R_1 + R_2 =I(\mathbf{U}_1;\mathbf{X}_1) + I(\mathbf{U}_2;\mathbf{X}_2) - I (\mathbf{U}_1; \mathbf{U}_2).
\label{eq:sum-rate-DIB-simplified-form}
\end{equation}
Similar in spirit to the IB method, the result of~\cite{aguerri21} shows that, in the distributed (or multi-view) case the encoders should learn representations that are \textit{maximally} informative about the label $\mathbf{Y}$ (in the sense of large $I(\mathbf{U}_1, \mathbf{U}_2; \mathbf{Y})$) while being of \textit{minimal} (sum) information comlexity as measured by~\eqref{eq:sum-rate-DIB-simplified-form}. Investigating the RHS of~\eqref{eq:sum-rate-DIB-simplified-form}, it is clear that this advocates in favor of large statistical correlations among the representations, i.e., large $I(\mathbf{U}_1;\mathbf{U}_2)$ -- Such correlations are enabled and learned during the training phase, e.g., through error-vector back-propagation as done in~\cite{aguerri21}. We hasten to mention, however, that the approaches of ~\cite{aguerri21} and this paper are different, the former being rate-distortion theoretic using a connection with a nultiterminal source coding problem while the latter is more \textit{direct } and statistical-learning oriented, obtained through direct bounding of the generalization error in terms of MDL. 

Theorem~\ref{th:distributed_gen} also sheds light on why the marginal-only regularizer of Section~\ref{sec:marginal_reg} is suboptimal: by treating each view independently, it ignores the negative (beneficial) joint-MDL term in~\eqref{eq:Dist_Redundancies2}, thereby penalizing cross-view redundancies that the bound actually rewards. Since the joint MDL term is difficult to estimate directly, the next section introduces a data-dependent Gaussian product mixture prior that captures the distributed structure and can be deployed efficiently.

\section{Gaussian Product Mixture Regularizer for Distributed Multiview Representation Learning} \label{sec:joint_multivew_regularizer}

Since small $\textnormal{MDL}(\vc{Q})$ implies good generalization (Theorems~\ref{th:generalizationExp_old} and~\ref{th:generalizationExp_reg}), it is natural to use it directly as a training regularizer. Doing so in the distributed multiview setting raises four concrete difficulties. \textbf{i.} To be effective, the prior must be data-dependent and updated throughout the training using statistics of latent variables \cite{dziugaite2018data,perez2021tighter,sefidgaran2023minimum}. \textbf{ii.} The joint latent dimension scales as $Kd$, making high-dimensional statistics increasingly hard to estimate reliably. \textbf{iii.} In contrast to the single-view case, inter-view correlations prevent the covariance of the joint latent vector from being diagonal. \textbf{iv.} Even if centralized joint statistics were available, the distributed architecture calls for local processing at each encoder rather than routing all computations to the server.

To address these four points simultaneously, we model, for each target class $c\in[C]$, the prior $Q_c$ as a \emph{Gaussian product mixture} whose parameters are updated incrementally from mini-batch statistics during training. Three properties justify this choice: \textbf{i.} it leads to an efficient distributed update protocol; \textbf{ii.} it penalizes cross-view redundancies less than marginal-only priors, in accordance with Theorem~\ref{th:distributed_gen}; \textbf{iii.} the marginal induced for each view $k$ is a Gaussian mixture, making the approach locally consistent with the single-view construction of \cite{sefidgaran2025generalization}. Property \textbf{(3)} is grounded in two facts: Gaussian mixtures are universal approximators of distributions \cite{dalal1983approximating,goodfellow2016deep,nguyen2022improving}, and among all distributions $q$, the minimizer of $\sum_{i\in[N]} D_{KL}(p_i\|q)$ is the mixture $q=\frac{1}{N}\sum_i p_i$, which is Gaussian when each $p_i$ is Gaussian.

In the following, we assume that $P_{\vc{U},\vc{U}'|\vc{X},\vc{X}',W_e}$ follows~\eqref{def:gaussian_encoders}. For every target $c\in[C]$, let $Q_c$ be defined over $\mathbb{R}^{Kd}$ and let $\vc{Q}(\vc{U},\vc{U}'|S,S',\hat{W}_e)=\prod_{i\in[n]} Q_{Y_i}(U_i)Q_{Y'_i}(U'_i)$. The regularizer then takes the form
\begin{equation}
    \textnormal{Regularizer}(\vc{Q}) = \sum\nolimits_{i\in[b]} D_{KL}\big(P_{U_i|x_i,w_e} \big\| Q_{y_i}(U_i)\big).
\end{equation}
As shown in \cite{sefidgaran2025generalization} and summarized in Appendix~\ref{sec:single_view}, a Gaussian mixture is a suitable prior for the single-view case or for treating each view marginally. Since, given the encoder $W_e$ and input $X=(X_1,\ldots,X_K)$, the latent variables $(U_1,\ldots,U_K)$ are conditionally independent, it is natural to model the joint prior for a given target as a mixture of Gaussian products. The lossless case is presented below; the extended approach using a more accurate KL-divergence estimate is detailed in Appendices~\ref{sec:lossless_multi_view} and~\ref{sec:lossy_multi_view}.

\begin{remark}[Instantiation for regression] \label{rem:regression_prior}
The construction above is written for the classification task, where the target-indexed family $\{Q_c\}_{c\in[C]}$ makes $\vc{Q}$ conditionally symmetric in the sense of Definition~\ref{def:symmetry}, as required by Theorem~\ref{th:generalizationExp_old}. Theorem~\ref{th:generalizationExp_reg} instead requires the fully symmetric prior of Definition~\ref{def:symmetry_reg}. For the regression tasks we therefore use a \emph{single} Gaussian product mixture $Q$, shared by all samples, \ie we set $\vc{Q}(\vc{U},\vc{U}'|S,S',\hat{W}_e)=\prod_{i\in[n]}Q(U_i)Q(U'_i)$; formally this amounts to taking $C=1$ in all the equations of this section and of Appendix~\ref{sec:average_estimate}, and all update rules, the distributed protocol, and the attention interpretation apply unchanged.
\end{remark}

\textbf{Lossless Gaussian product mixture.}
For every $c\in[C]$, the prior $Q_c$ over $\mathbb{R}^{Kd}$ is defined as
\begin{align}
   Q_{c} = \sum\nolimits_{\vc{m}\in[M]^K} \alpha_{c,\vc{m}}\, Q_{c,\vc{m}},
\end{align}
where $\vc{m}=(m_1,\ldots,m_K)\in[M]^K$, $\alpha_{c,\vc{m}}\in[0,1]$ with $\sum_{\vc{m}\in[M]^K}\alpha_{c,\vc{m}}=1$, and each component is a product of $K$ marginal Gaussians:
\begin{align*}
Q_{c,\vc{m}}= \prod\nolimits_{k\in[K]} Q_{c,k,m_k}, \qquad Q_{c,k,m}=\mathcal{N}\left(\mu_{c,k,m},\diag\left(\sigma_{c,k,m}^2\right)\right).
\end{align*}
Defining the marginal coefficients $\alpha_{c,k,m} = \sum_{\vc{m}\in[M]^K\colon m_k=m}\alpha_{c,\vc{m}}$, the marginal prior of view $k$ under $Q_c$ is $Q_{c,k} = \sum_{m\in[M]}\alpha_{c,k,m}\,Q_{c,k,m}$, which is itself a Gaussian mixture. This joint prior thus induces marginal Gaussian mixture priors for each view, consistent with \cite{sefidgaran2025generalization}. The regularizer restricted to mini-batch $\mathcal{B}$ is $\textnormal{Regularizer}(\vc{Q})\coloneqq D_{KL}\big(P_{\vc{U}_{\mathcal{B}}|\vc{X}_{\mathcal{B}},W_e}\big\|\vc{Q}_{\mathcal{B}}\big)$.

\textbf{Initialization.} The priors $Q_c^{(0)}$ are initialized via a novel distributed variant of k-means$++$ \cite{arthur2007k}, described in Appendix~\ref{sec:initialization_multi_view}.

\textbf{Update of the priors.} In iteration $(t)$, the regularizer admits the variational upper bound
\begin{align}
    &\textnormal{Regularizer}(\vc{Q})
     \leq\nonumber \\
     &\sum\nolimits_{i\in[b]} \sum\nolimits_{\vc{m}\in[M]^K} \gamma_{i,\vc{m}} \Big(\sum     \nolimits_{k\in[K]}D_{KL}\big(P_{U_{i,k}|x_{i,k},w_{e,k}} \big\| Q^{(t)}_{y_i,k,m_k}(U_{i,k})\big)-\log\big(\alpha_{y_i,\vc{m}}^{(t)}/\gamma_{i,\vc{m}}\big)\Big),  \nonumber
\end{align}
for any $\gamma_{i,\vc{m}}\geq 0$ with $\sum_{\vc{m}\in[M]^K}\gamma_{i,\vc{m}}=1$ for every $i\in[b]$. The minimizing coefficients are
\begin{align}
    \gamma_{i,\vc{m}} = \frac{\alpha_{y_i,\vc{m}}^{(t)} e^{-\sum_{k\in[K]}D_{KL}\big(P_{U_{i,k}|x_{i,k},w_{e,k}} \big\| Q^{(t)}_{y_i,k,m_k}(U_{i,k})\big)}}{\sum_{\vc{m}'\in [M]^K} \alpha_{y_i,\vc{m}'}^{(t)} e^{-\sum_{k\in[K]}D_{KL}\big(P_{U_{i,k}|x_{i,k},w_{e,k}} \big\| Q^{(t)}_{y_i,k,m'_k}(U_{i,k})\big)}}, \quad \vc{m}\in[M]^K. \label{eq:gamma_i_dist}
\end{align}
Denoting the marginals as $\gamma_{i,k,m}=\sum_{\vc{m}\in[M]^K\colon m_k=m}\gamma_{i,\vc{m}}$ and defining $\gamma_{i,c,\vc{m}}$, $\gamma_{i,c,k,m}$ accordingly, the optimal joint coefficients are
\begin{align*}
    \alpha_{c,\vc{m}}^* = b_{c,\vc{m}}/b_c, \qquad b_{c,\vc{m}}=\sum\nolimits_{i\in[b]}\gamma_{i,c,\vc{m}}, \qquad b_c=\sum\nolimits_{\vc{m}\in[M]^K}b_{c,\vc{m}},
\end{align*}
updated as $\alpha_{c,\vc{m}}^{(t+1)}=(1-\eta_3)\alpha_{c,\vc{m}}^{(t)}+\eta_3\alpha_{c,\vc{m}}^*$. The marginal parameters $\mu_{c,k,m}$ and $\sigma_{c,k,m}$ of each view are then updated locally using the marginal coefficients $\gamma_{i,k,m}$. Due to lack of space, the details are provided in Appendix~\ref{sec:lossless_multi_view} (see steps \eqref{eq:optimal_updates_lossless_multi} and \eqref{eq:updates_lossless_multi_app}).

\textbf{Regularizer.} Using \eqref{eq:gamma_i_dist}, the regularizer simplifies to
\begin{align}
    - \sum\nolimits_{i\in[b]} \log\left(\sum\nolimits_{\vc{m}\in [M]^K} \alpha_{y_i,\vc{m}}^{(t)} e^{-\sum_{k\in[K]}D_{KL}\big(P_{U_{i,k}|x_{i,k},w_{e,k}}\|Q_{y_i,k,m_k}^{(t)}\big)} \right).\label{eq:reg_lossless_multi}
\end{align}

\textbf{Distributed implementation.} As detailed in Appendix~\ref{sec:lossless_multi_view} (see~\eqref{eq:optimal_updates_lossless_multi}), the per-view parameters $\mu_{c,k,m}^*$ and $\sigma_{c,k,m}^*$ depend only on the marginal coefficients $\gamma_{i,c,k,m}$ and the $k$-th view's latent variables. The distributed update protocol proceeds as follows:
\begin{itemize}[leftmargin=*]
    \item Each client $k$ shares $D_{KL}\big(P_{U_{i,k}|x_{i,k},w_{e,k}}\big\|Q^{(t)}_{y_i,k,m_k}(U_{i,k})\big)$ for $i\in[b]$ and $m\in[M]$.
    \item The server updates the joint coefficients $\alpha_{c,\vc{m}}^{(t+1)}$, sends the marginals $\alpha_{c,k,m}^{(t+1)}$ to each client $k$, computes the regularization term, performs the forward pass and backpropagation, and returns the gradient vectors to the clients.
    \item Each client updates its local marginal prior and encoder using the received gradients and marginal coefficients $\alpha_{c,k,m}^{(t+1)}$.
\end{itemize}
This protocol exploits distributed computational resources and keeps communication overhead minimal.

\textbf{Penalizing redundancies.} As shown in Appendix~\ref{sec:redundancyExample}, the regularizer~\eqref{eq:reg_lossless_multi} penalizes cross-view redundancies in the latent variables less than the marginal-only regularizer of Section~\ref{sec:marginal_reg}, making it a more suitable choice in light of Theorem~\ref{th:distributed_gen}.

\textbf{Lossy Gaussian product mixture.} The lossy extension is detailed in Appendix~\ref{sec:lossy_multi_view}. In the lossy case, the coefficient $\gamma_{i,\vc{m}}$ takes the form
\begin{equation*}
\gamma_{i,\vc{m}} \propto \beta_{y_i,\vc{m}}^{(t)}\exp\left(\frac{\sum_{k\in[K]}\langle\mu_{x_i,k},\mu_{y_i,k,m_k}\rangle}{\sqrt{Kd}}\right),
\end{equation*}
where $\beta_{i,\vc{m}}^{(t)}\propto\alpha_{i,\vc{m}}^{(t)}$ when component means are normalized. This update rule can be interpreted as a \emph{weighted distributed attention mechanism}: the contribution of each joint component $Q_{c,\vc{m}}$ is determined by how much its $K$ marginal components jointly ``attend'' to the corresponding latent variables $\{U_{i,1},\ldots,U_{i,K}\}$.

\section{Experiments} \label{sec:experiments}

In this section, we present the results of our simulations. The reader is referred to Appendix~\ref{sec:details_exp} for additional details, including used datasets, models, training hyperparameters, and more results. 

In the experiments, we considered the lossy regularizer approach with the Gaussian product mixture prior and the KL-divergence estimate of $\left(D_{\text{prod}}+D_{\text{var}}\right)/2$, as detailed in Appendix~\ref{sec:average_estimate}. In this section, we refer to our regularizer as \emph{Gaussian product mixture MDL} (GPM-MDL). To verify the practical benefits of the introduced regularizer, we conducted several experiments considering different datasets and encoder architectures as summarized below and detailed in Appendix~\ref{sec:details_exp}. The considered number of views $K$ are 2, 3, 4, and 8. The encoder architectures include CNN4 and ResNet18. The tested datasets are CIFAR10, CIFAR100 for image classification, and IMDB-WIKI for face age prediction.

For the multiview setup, the inputs of each encoder were copies of the same image with independent distortion. Several methods of generating multiview data have been considered, including: \textbf{i.} Adding independently to each view one of the three different levels of erasure rates and random transformations, called Light, Medium, and Heavy, as detailed in Table~\ref{table:noise_levels} of Appendix~\ref{experiments:additional_results}. \textbf{ii.} Occluding the image so that each view observes only a portion of the image: This includes splitting the image into the left (L) and right (R) parts or upper (U) and bottom (B) parts, with small overlaps, or a combination of them. \textbf{iii.} A combination of the two above methods. In particular, we considered several setups where the qualities of the views are similar, as well as numerous setups where the views are unevenly informative. 

We compared our approach with the no-regularizer case, as well as with the case where per-view VIB regularizer is applied. The results of the extensive simulations reported in Table~\ref{table:experiments_multiview} validate the benefit of using GMP-MDL regularizer. Classification and regression experiments are run independently for 5 times. In Table~\ref{table:experiments_multiview}, we reported for each regularizer the best achieved average test accuracy, for the different tested trade-off regularization parameter.

\begin{table}[h] \renewcommand{\arraystretch}{1.5}
\hspace{.1 cm}    
   \begin{tabular}{|c|c|l|c|c|c|}
           \hline 
           \# & K & \hspace{3.8 cm} Scenario &  no reg. & VIB & \textbf{GPM-MDL} \\ \hline
           1 & 2&CIFAR10, Enc.=CNN4, Dist.=(Light,Light) &  63.2\% & 63.9\% & \textbf{67.5\%}\\
           \hline
           2 &2& CIFAR100, Enc.=ResNet18, Dist.=(Light,Light) &  42.6\% & 44.1\% & \textbf{46.8\%}\\
           \hline 
            3 & 2&CIFAR10, Enc.=CNN4, Dist.=Occ.(L,R)  &  60.7\% & 61.0\% & \textbf{65.2\%}\\
             \hline
           4 & 2&CIFAR10, Enc.=CNN4, Dist.=  Occ.(L,R) 
 + (Light, Light) &  56.0\% & 56.7\% & \textbf{60.6\%}\\
           \hline
             5 & 3 & CIFAR100, Enc.=ResNet18, Dist.=(Medium, Medium, Medium) &  37.3\% & 38.1\% & \textbf{41.2\%}\\
            \hline
           6 & 4 & CIFAR10, Enc.=CNN4, Dist.= Occ.(L,R,U,B) & 64.6\% & 64.7\% & \textbf{67.4\%}\\
           \hline
           7 & 4 & CIFAR10, Enc.=CNN4, Dist.=Occ.(L,R,U,B)+(Light,\ldots,Light) & 59.9\% & 60.1\% & \textbf{63.4\%}\\
           \hline 
           8 & 8 & CIFAR10, Enc.=CNN4, Dist.=(Heavy,Heavy,\ldots,Heavy) & 39.6\% & 44.7\% & \textbf{52.9\%}\\ \hline
             \hline 9 & 2 & IMDB-WIKI, Enc.=ResNet18,  Dist.= (Light,Light) & 0.487
 & 0.486 & \textbf{0.478}\\
           \hline 10 & 2 & IMDB-WIKI, Enc.=ResNet18, Dist.=(Medium,Medium) & 0.492 & 0.491 & \textbf{0.485}\\
            \hline
        \end{tabular}\vspace{0.2 cm}
    \caption{Test performance of multiview representation learning models across encoder architectures, datasets, and regularizers: none, per-view VIB \cite{alemi2016deep}, and our GPM-MDL. Classification accuracy in [\%]; regression accuracy as mean squared prediction error of normalized age labels. Distortion levels defined in Table~\ref{table:noise_levels} (Appendix~\ref{experiments:additional_results}). Enc. = Encoder, Dist. = Distortion, Occ. = Occlusion.}
    \label{table:experiments_multiview}
\end{table}


\section{Proof of Theorem~\ref{th:generalizationExp_reg}} \label{pr:generalizationExp_reg}

Before entering the details, we recall the structure of the argument, sketched in Section~\ref{sec:Bounds_multi_view}. The expected generalization error is first split, via the ghost dataset, into a term $\Delta_1$ measuring a train/test asymmetry of the latent variables at a common target, and a term $\Delta_2$ measuring the alignment between the targets and the predictions formed at the same sample index. Each term is then handled by a change of measure to the prior $\vc{Q}$, which produces $\textnormal{MDL}(\vc{Q})$ plus a cumulant generating function that has to be bounded under $\vc{Q}$. The symmetry of $\vc{Q}$ is what allows a randomization to be injected for free at this stage: in both cases the targets are held fixed and the latent variables are randomized, but along different axes: independent sign flips of the pairs $(U_i,U'_i)$ for $\Delta_1$, and a uniformly random permutation of the training indices for $\Delta_2$. Each randomization is chosen so that the quantity to be bounded has zero mean under it. The first yields independent summands and is handled by Hoeffding's lemma; the second yields \emph{dependent} terms (sampling without replacement) and is handled by Stein's method for exchangeable pairs, in Lemma~\ref{lem:stein_bound} below. Both give the same sub-Gaussian proxy $R^2/n$, and a single optimization over the change-of-measure parameter $\lambda$ concludes.

 Denote the prediction of the model $W$ for training input $X_i$ by $\hat{Y}_i$ and for the test input $X'_i$ by $\hat{Y}'_i$. We have
\begin{align}
    \gen_r(S,W) = \E_{S'} \E_{\vc{U},\vc{U}',\hat{\vc{Y}},\hat{\vc{Y}}'\sim P_1}    \left[\frac{1}{n}\sum_{i=1}^n \left( \|Y'_i - \hat{Y}'_i\|^2 -\|Y_i - \hat{Y}_i\|^2 \right)\right].
\end{align}
where 
\begin{align}
    P_1 \triangleq P_{\vc{U},\vc{U}'|\vc{X},\vc{X}',W_e} P_{\hat{\vc{Y}},\hat{\vc{Y}}'|\vc{U},\vc{U}',W_d}.
\end{align}
Denote also
\begin{align}
     P_2 \triangleq \vc{Q}(\vc{U},\vc{U'}|S,S',W_e) P_{\hat{\vc{Y}},\hat{\vc{Y}}'|\vc{U},\vc{U}',W_d}.
\end{align}

We borrow the decomposition of \cite{futami2025information}; but analyze each term differently than the approach of \cite{futami2025information} that relies on the discrete mapping of the latent variables in the `VQ' step.

Define $\ell_{a} \colon \mathcal{Y} \times \hat{\mathcal{Y}} \times \hat{\mathcal{Y}} \to \R$ and  $\ell_{b} \colon \hat{\mathcal{Y}}  \times  \mathcal{Y} \times  \mathcal{Y} \to \R$ as
\begin{align}
\ell_{a}(y',\hat{y}',\hat{y}) =& \big\| \hat{y}'- y'\big\|^2-\big\| \hat{y}- y'\big\|^2 \nonumber \\
\ell_{b}(\hat{y},y',y) =&  \big\| \hat{y} - y'\big\|^2-\big\| \hat{y} - y\big\|^2. \nonumber
\end{align}
With this definition, we decompose the expected generalization error as
\begin{align}
    \E_{S,W}\left[\gen_r(S,W)\right] =& \E_{S,W,S'} \E_{\vc{U},\vc{U}',\hat{\vc{Y}},\hat{\vc{Y}}'\sim P_1 }    \left[\frac{1}{n}\sum_{i=1}^n \ell_{a}(Y'_i,\hat{Y}'_i,\hat{Y}_i) \right] \label{eq:term_1}\\
    &+\E_{S,W,S'} \E_{\vc{U},\vc{U}',\hat{\vc{Y}},\hat{\vc{Y}}'\sim P_1 }    \left[\frac{1}{n}\sum_{i=1}^n \ell_{b}(\hat{Y}_i,Y'_i,Y_i) \right] \label{eq:term_2}.
\end{align}
We bound each of the two terms in \eqref{eq:term_1} and \eqref{eq:term_2}, denoted respectively as $\Delta_1$ and $\Delta_2$ separately. These two terms play complementary roles. In $\Delta_1$, the \emph{same} target $Y'_i$ is used to score both the test prediction $\hat{Y}'_i$ and the training prediction $\hat{Y}_i$; the term thus measures a train/test asymmetry carried by the latent pair $(U_i,U'_i)$, and it vanishes in expectation as soon as the two elements of that pair are exchanged at random, by the antisymmetry established in \eqref{eq:delta_ell_bound}. In $\Delta_2$, the \emph{same} prediction $\hat{Y}_i$ is scored against the training target $Y_i$ and against the independent copy $Y'_i$; the term thus measures the alignment between a target and the prediction formed at the \emph{same} sample index, and it vanishes in expectation as soon as that index matching is destroyed by a random relabelling of the training indices (see \eqref{eq:C_centered}). In both cases the targets are held fixed and it is the latent variables that are randomized; what differs is the axis along which the randomization acts. Accordingly, the symmetrization used for $\Delta_1$ acts on the train/test axis, whereas the one used for $\Delta_2$ acts on the sample axis.

\textbf{Bounding \eqref{eq:term_1}.}
Denote
\begin{align}
    \ell_{a}(U'_i,U_i|Y'_i) \triangleq & \E_{\hat{Y}'_i,\hat{Y}_i  \sim P_{\hat{Y}'_i,\hat{Y}_i|U'_i,U_i,W_d}}[\ell_{a}(Y'_i,\hat{Y}'_i,\hat{Y}_i)] \nonumber \\
    = &\E_{\hat{Y}'_i \sim P_{\hat{Y}'_i|U'_i,W_d}}[\ell_{r}(Y'_i,\hat{Y}'_i)] -  \E_{\hat{Y}_i \sim P_{\hat{Y}_i|U_i,W_d}}[\ell_{r}(Y'_i,\hat{Y}_i)].
\end{align}
Note that
\begin{align}
    \ell_{a}(U'_i,U_i|Y'_i) =& - \ell_{a}(U_i,U'_i|Y'_i), \nonumber \\
    |\ell_{a}(U'_i,U_i|Y'_i)| \leq  & R, \label{eq:delta_ell_bound}
\end{align}
where the first identity expresses the \emph{antisymmetry} of $\ell_a$ in its two latent arguments, and the second inequality comes from the fact that $|\ell_{a}(U'_i,U_i|Y'_i)|$ is less than or equal to $\E_{\hat{Y}'_i,\hat{Y}_i  \sim P_{\hat{Y}'_i,\hat{Y}_i|U'_i,U_i,W_d}}[|\ell_{a}(Y'_i,\hat{Y}'_i,\hat{Y}_i)|]$ and
\begin{align}
|\ell_{a}(Y'_i,\hat{Y}'_i,\hat{Y}_i)| \leq &  \max\left(\big\| \hat{Y}'_i- Y'_i\big \|^2,\big\| \hat{Y}_i- Y'_i\big \|^2\right) \leq R,
\end{align}
using Assumption~\ref{assum:bounded}, since $\hat{\mathcal{Y}}=\mathcal{Y}$. These two properties of $\ell_a$, \ie antisymmetry and boundedness by $R$, are all that the next step uses.

For any $\lambda \in \R$, applying Donsker-Varadhan's inequality for changing measure from $P_{\vc{U},\vc{U}'|\vc{X},\vc{X}',W_e}$ to the measure $\vc{Q}(\vc{U},\vc{U'}|S,S',W_e)$ gives
\begin{align}
   \lambda \Delta_1 \leq  \textnormal{MDL}(\vc{Q}) + \E_{S,S',W_e} \left[\log \E_{\vc{U},\vc{U}'\sim \vc{Q}} \exp\left(\frac{\lambda}{n}\sum_{i=1}^n  \ell_{a}(U'_i,U_i|Y'_i)) \right)\right] \label{eq:dv_term_1}
\end{align}
where $ \textnormal{MDL}(\vc{Q}) \coloneqq   \E_{S,S',W_e} \big[ D_{KL}\big(P_{\vc{U}, \vc{U}'|\vc{X},\vc{X}',W_e} \big\| \vc{Q} \
     \big) \big]$. 

Now to bound $\E_{\vc{U},\vc{U}'\sim \vc{Q}} \exp\left(\frac{\lambda}{n}\sum_{i=1}^n \ell_{a}(U'_i,U_i|Y'_i) \right)$, we fix the set of values $(\vc{U},\vc{U}')$ and consider any permutation of them, exploiting the fact that $\vc{Q}$ is symmetric under any permutation. More precisely, Let $\vc{J}=(J_1,\ldots,J_n)$ where $J_i$ are distributed i.i.d and uniformly over $\{1,2\}$. Let $\tilde{\vc{U}}\in \mathcal{U}^{n\times 2}$ be defined as following. $\tilde{U}_{i,J_i} = U_i$ and $\tilde{U}_{i,J_i^c} = U'_i$, where $J_i^c = 3-J_i$. In words, $\vc{J}$ decides, independently for each index $i$, which element of the pair $(U_i,U'_i)$ is declared to be the ``training'' one. Since $\vc{Q}$ is symmetric in the sense of Definition~\ref{def:symmetry_reg}, and the transpositions $(i,i+n)$, $i\in[n]$, are particular permutations of $[2n]$, the law of $\tilde{\vc{U}}$ under $\vc{Q}\otimes\unif(\{1,2\})^{\otimes n}$ coincides with that of $(\vc{U},\vc{U}')$ under $\vc{Q}$; the randomization is therefore free of charge. Note that this is the only place where the symmetry of $\vc{Q}$ is used in bounding $\Delta_1$.

Thus
\begin{align}
    \E_{\vc{U},\vc{U}'\sim \vc{Q}} \exp&\left(\frac{\lambda}{n}\sum_{i=1}^n  \ell_{a}(U'_i,U_i|Y'_i) ) \right) \nonumber \\
    =&  \E_{\vc{U},\vc{U}'\sim \vc{Q}} \E_{\vc{J}\sim \text{Unif}(\{1,2\})^{\otimes n}} \exp\left(\frac{\lambda}{n}\sum_{i=1}^n \ell_{a}(\tilde{U}_{i,1},\tilde{U}_{i,2}|Y'_i) \right) \nonumber \\
   =&  \E_{\vc{U},\vc{U}'\sim \vc{Q}} \E_{\vc{J}\sim \text{Unif}(\{1,2\})^{\otimes n}} \exp\left(\frac{\lambda}{n}\sum_{i=1}^n  (-1)^{J_i}\ell_{a}(U'_i,U_i|Y'_i) \right) \nonumber \\
   \stackrel{(a)}{\leq} &  \E_{\vc{U},\vc{U}'\sim \vc{Q}} \exp\left(\frac{ \lambda^2 R^2 }{2n} \right) \nonumber \\
   = &   \exp\left(\frac{ \lambda^2 R^2 }{2n} \right) \label{eq:dv_term_1_cgf}
\end{align}
where the second equality uses the antisymmetry in \eqref{eq:delta_ell_bound}, and $(a)$ is derived using Hoeffding's lemma, applied conditionally on $(\vc{U},\vc{U}')$ to the independent, centered random variables $(-1)^{J_i}\ell_{a}(U'_i,U_i|Y'_i)$, each supported on an interval of length at most $2R$ since $|\ell_{a}(\tilde{U}_{i,1},\tilde{U}_{i,2}|Y'_i)| \leq  R$ by \eqref{eq:delta_ell_bound}.

Plugging \eqref{eq:dv_term_1_cgf} into \eqref{eq:dv_term_1} gives
\begin{align}
   \Delta_1 \leq  \frac{1}{\lambda} \textnormal{MDL}(\vc{Q}) + \frac{ \lambda R^2 }{2n}. \label{eq:dv_term_1_final}
\end{align}

\textbf{Bounding \eqref{eq:term_2}.} Denote
\begin{align}
    g(u,w_d) \triangleq & \E_{\hat{Y} \sim P_{\hat{Y}|u,w_d}}[\hat{Y}_i], \nonumber \\
    \bar{g}(\vc{U},w_d) = & \frac{1}{n}\sum_{i=1}^n g(U_i,w_d).
\end{align}
Since marginal distributions of $\vc{Y}$ and $\vc{Y}'$ are the same; by expanding the squared error loss, we have
\begin{align}
    \Delta_2 = & \E_{S,W,S'} \E_{\vc{U},\vc{U}',\hat{\vc{Y}},\hat{\vc{Y}}'\sim P_1}    \left[\frac{1}{n}\sum_{i=1}^n  \ell_{b}(\hat{Y}_i,Y'_i,Y_i)\right] \nonumber \\
    = & \E_{S,W} \E_{\vc{U},\hat{\vc{Y}}\sim P_1}    \left[ \frac{2}{n}\sum_{i=1}^n  \left\langle Y_i - \E_{Y}[Y], \hat{Y}_i \right\rangle \right]\nonumber \\
    = & \E_{S,W,S'} \E_{\vc{U}\sim P_{\vc{U}|\vc{X},W_e}}    \left[ \frac{2}{n}\sum_{i=1}^n  \left\langle  Y_i - \E_{Y}[Y],\E_{\hat{Y}_i \sim P_{\hat{Y}_i|U_i,W_d}}[\hat{Y}_i]\right\rangle \right]\nonumber \\
    = & \E_{S,W,S'} \E_{\vc{U}\sim P_{\vc{U}|\vc{X},W_e}}    \left[ \frac{2}{n}\sum_{i=1}^n  \left\langle  Y_i - \E_{Y}[Y],g(U_i,W_d)\right\rangle \right]\nonumber \\
    = & \E_{S,W,S'} \E_{\vc{U}\sim P_{\vc{U}|\vc{X},W_e}}    \left[ \frac{2}{n}\sum_{i=1}^n  \left\langle  Y_i - \E_{Y}[Y],\bar{g}(\vc{U},W_d)\right\rangle \right]\nonumber \\
    & + \E_{S,W} \E_{\vc{U}\sim P_{\vc{U}|\vc{X},W_e}}    \left[ \frac{2}{n}\sum_{i=1}^n  \left\langle  Y_i - \E_{Y}[Y],g(U_i,W_d)-\bar{g}(\vc{U},W_d)\right\rangle \right]. \label{eq:term2_decomposed}
\end{align}
We start by bounding the first term.
\begin{align}
    \E_{S,W,S'} \E_{\vc{U}\sim P_{\vc{U}|\vc{X},W_e}}   & \left[ \frac{2}{n}\sum_{i=1}^n  \left\langle  Y_i - \E_{Y}[Y],\bar{g}(\vc{U},W_d)\right\rangle \right]  \nonumber \\
    \stackrel{(a)}{=} & \E_{S,W,S'} \E_{\vc{U}\sim P_{\vc{U}|\vc{X},W_e}}    \left[ \frac{2}{n}\sum_{i=1}^n  \left\langle  Y_i - \E_{Y}[Y],\bar{g}(\vc{U},W_d)- \E_{Y}[Y]\right\rangle \right]  \nonumber \\
    = & \E_{S,W,S'} \E_{\vc{U}\sim P_{\vc{U}|\vc{X},W_e}} \left[ 2  \left\langle \frac{1}{n}\sum_{i=1}^n (Y_i - \E_{Y}[Y]),\bar{g}(\vc{U},W_d)- \E_{Y}[Y]\right\rangle \right]  \nonumber \\
     \stackrel{(b)}{\leq} & \E_{S,W,S'} \E_{\vc{U}\sim P_{\vc{U}|\vc{X},W_e}} \left[ 2  \left\|\frac{1}{n}\sum_{i=1}^n (Y_i - \E_{Y}[Y])\right\|\, \left\|\bar{g}(\vc{U},W_d)- \E_{Y}[Y]\right\|\right] \nonumber\\
     \stackrel{(c)}{\leq} & 2\sqrt{R}\E_{\vc{Y}} \left[ \left\|\frac{1}{n}\sum_{i=1}^n (Y_i - \E_{Y}[Y])\right\|\right] \nonumber\\
     \stackrel{(d)}{\leq} & 2\sqrt{R}\E_{\vc{Y}} \left[ \left\|\frac{1}{n}\sum_{i=1}^n (Y_i - \E_{Y}[Y])\right\|^2\right]^{1/2} \nonumber\nonumber\\
     \stackrel{(e)}{\leq} & \frac{2\sqrt{R}}{\sqrt{n}} \text{Var}(Y)^{1/2}\nonumber\nonumber\\
     \stackrel{(f)}{\leq} & \frac{R\sqrt{2}}{\sqrt{n}},  \label{eq:dv_term_2_exp_dec}
\end{align}
where $(a)$ follows from a simple algebra, $(b)$ is derived using Cauchy-Schwarz inequality, $(c)$ by Assumption~\ref{assum:bounded}, $(d)$ by Jensen Inequality, $(e)$ by independence of $Y_i$ and simple algebra, $(f)$ by noting that $Var(Y) = \frac{1}{2}\E_{Y,Y'}\big[\|Y-Y'\|^2\big]$ and using Assumption~\ref{assum:bounded}.

Plugging \eqref{eq:dv_term_2_exp_dec} into \eqref{eq:term2_decomposed} gives
\begin{align}
    \Delta_2 \leq & \frac{R\sqrt{2}}{\sqrt{n}} + \E_{S,W} \E_{\vc{U}\sim P_{\vc{U}|\vc{X},W_e}}    \left[ \frac{2}{n}\sum_{i=1}^n  \left\langle  Y_i - \E_{Y}[Y],g(U_i,W_d)-\bar{g}(\vc{U},W_d)\right\rangle \right]\nonumber \\
    = &\frac{R\sqrt{2}}{\sqrt{n}} + \E_{S,W,S'} \E_{\vc{U},\vc{U}'\sim P_{\vc{U}|\vc{X},\vc{X}',W_e}}    \left[ \frac{2}{n}\sum_{i=1}^n  \left\langle  Y_i - \E_{Y}[Y],g(U_i,W_d)-\bar{g}(\vc{U},W_d)\right\rangle \right]. \label{eq:term2_decomposed_2}
\end{align}
Now, for any $\lambda \in \R$, applying Donsker-Varadhan's inequality for changing measure from $P_{\vc{U},\vc{U}'|\vc{X},\vc{X}',W_e}$ to $\vc{Q}(\vc{U},\vc{U'}|S,S',W_e)$ gives

\begin{align}
   \lambda \Delta_2 \leq & \frac{\lambda R\sqrt{2}}{\sqrt{n}} + \E_{S,W,S'} \E_{\vc{U}\sim P_{\vc{U},\vc{U}'|\vc{X},\vc{X}',W_e}}    \left[ \frac{2\lambda}{n}\sum_{i=1}^n  \left\langle  Y_i - \E_{Y}[Y],g(U_i,W_d)-\bar{g}(\vc{U},W_d)\right\rangle \right]\nonumber \\
   \leq & \frac{\lambda R\sqrt{2}}{\sqrt{n}} +   
   \textnormal{MDL}(\vc{Q}) \nonumber \\
   &+ \E_{S,S',W_e} \left[\log \E_{\vc{U},\vc{U}'\sim \vc{Q}} \exp\left(\frac{2\lambda}{n}\sum_{i=1}^n  \left\langle  Y_i - \E_{Y}[Y], g(U_i,W_d)-\bar{g}(\vc{U},W_d)\right\rangle \right)\right], \label{eq:dv_term_2}
\end{align}
where $ \textnormal{MDL}(\vc{Q}) \coloneqq   \E_{S,S',W_e} \big[ D_{KL}\big(P_{\vc{U}, \vc{U}'|\vc{X},\vc{X}',W_e} \big\| \vc{Q} \
     \big) \big]$. 
Denote
\begin{align}
    \widebar{\vc{Y}} =& \frac{1}{n}\sum_{i=1}^n Y_i,\nonumber\\
    V_i \triangleq & Y_i-\widebar{\vc{Y}}.
\end{align}
Note that for every $i\in[n]$ , using simple algebra and using Assumption~\ref{assum:bounded}, we have
\begin{align}
 \sum_{i=1}^n \|V_i\|^2 = \frac{1}{2n} \sum_{i=1}^n\sum_{j=1}^n  \|Y_i-Y_j\|^2 \leq \frac{nR}{2}.     \label{eq:v_bound}
\end{align}
With this notation
\begin{align}
   \frac{1}{n}\sum_{i=1}^n & \left\langle  Y_i - \E_{Y}[Y], g(U_i,W_d)-\bar{g}(\vc{U},W_d)\right\rangle  \nonumber \\
    = & \frac{1}{n}\sum_{i=1}^n  \left\langle  \widebar{\vc{Y}}- \E_{Y}[Y], g(U_i,W_d)-\bar{g}(\vc{U},W_d)\right\rangle + \frac{1}{n}\sum_{i=1}^n  \left\langle  V_i, g(U_i,W_d)-\bar{g}(\vc{U},W_d)\right\rangle \nonumber \\
    = &   \left\langle  \widebar{\vc{Y}}- \E_{Y}[Y], \frac{1}{n}\sum_{i=1}^n  \left(g(U_i,W_d)-\bar{g}(\vc{U},W_d)\right)\right\rangle + \frac{1}{n}\sum_{i=1}^n  \left\langle  V_i, g(U_i,W_d)-\bar{g}(\vc{U},W_d)\right\rangle \nonumber \\
     = & \frac{1}{n}\sum_{i=1}^n  \left\langle  V_i, g(U_i,W_d)-\bar{g}(\vc{U},W_d)\right\rangle, \label{eq:dv_term_2_exp_dec_2}
\end{align}
using the definition of $\bar{g}(\vc{U},W_d)$. Combining this with \eqref{eq:dv_term_2} gives 
\begin{align}
   \lambda \Delta_2 
   \leq & \frac{\lambda R\sqrt{2}}{\sqrt{n}} +   
   \textnormal{MDL}(\vc{Q}) + \E_{S,S',W_e} \left[\log \E_{\vc{U},\vc{U}'\sim \vc{Q}} \exp\left(\frac{2\lambda}{n}\sum_{i=1}^n  \left\langle  V_i, g(U_i,W_d)-\bar{g}(\vc{U},W_d)\right\rangle \right)\right], \label{eq:dv_term_2_simp_1}
\end{align}
Hence, it remains to upper bound the last term in \eqref{eq:dv_term_2_simp_1}. We bound in particular $$\E_{\vc{U},\vc{U}'\sim \vc{Q}} \exp\left(\frac{2\lambda}{n}\sum_{i=1}^n  \left\langle  V_i, g(U_i,W_d)-\bar{g}(\vc{U},W_d) \right\rangle \right).$$ To bound this expectation, we fix the set of values in vectors $\vc{U},\vc{U}'$ and since $\vc{Q}$ is symmetric, we consider the expectation with respect to the uniform permutations of sample indices. More formally, let 
\begin{align}
    C_i \triangleq g(U_i,W_d)-\bar{g}(\vc{U},W_d).
\end{align}

Two centerings have now been performed which are both essential for what follows: the targets have been centered by their empirical mean, giving $\sum_{i\in[n]}V_i=0$, and the decoder outputs have been centered by their empirical mean, giving
\begin{align}
    \sum\nolimits_{i\in[n]}C_i = 0. \label{eq:C_centered}
\end{align}
Moreover, by Assumption~\ref{assum:bounded}, $\|V_i-V_j\|^2 = \|Y_i-Y_j\|^2\leq R$, and, since $g(u,w_d)$ is an average of elements of $\hat{\mathcal{Y}}=\mathcal{Y}$ and the diameter of a set equals that of its convex hull, also $\|C_i-C_j\|^2 = \|g(U_i,W_d)-g(U_j,W_d)\|^2\leq R$.

Let $\pi\colon [n] \to [n]$ be a permutation chosen uniformly over the set of all permutations from $[n]$ to $[n]$. Then
\begin{align}
     \E_{\vc{U},\vc{U}'\sim \vc{Q}} \exp\left(\frac{2\lambda}{n}\sum_{i=1}^n  \left\langle  V_i, g(U_i,W_d)-\bar{g}(\vc{U},W_d)\right\rangle \right) = \E_{\vc{U},\vc{U}'}\E_{\pi} \exp\left(\frac{2\lambda}{n}\sum_{i=1}^n  \left\langle  V_i, C_{\pi(i)}\right\rangle \right). \label{eq:perm_injection}
\end{align}

Identity \eqref{eq:perm_injection} holds because $\vc{Q}$ is symmetric: relabelling the training indices leaves the law of $\vc{U}$ unchanged, while the targets, and hence the vector $(V_1,\ldots,V_n)$, are held fixed. It is exactly here that the strengthening from the conditionally symmetric prior of Definition~\ref{def:symmetry} to the symmetric prior of Definition~\ref{def:symmetry_reg} is needed, since $\pi$ is not required to preserve the targets.

Unlike the sign randomization used for $\Delta_1$, the injected permutation does \emph{not} render the summands independent: a uniform permutation of $[n]$ samples indices \emph{without} replacement, so $\{\langle V_i,C_{\pi(i)}\rangle\}_{i\in[n]}$ is an exchangeable but strongly dependent family and Hoeffding's lemma no longer applies. The bounded-difference inequality does not give the correct dependence on $n$ either, since a single transposition can move the statistic by an amount of order $R$. For discrete latent variables one may circumvent this by partitioning the indices according to the codeword to which they are mapped, which is the route taken in \cite{futami2025information}; this is unavailable here, as we allow $\mathcal{U}=\R^d$. We instead use Stein's method for exchangeable pairs, developed by Chatterjee \cite{chatterjee2006stein}, which is tailored to precisely such permutation statistics. For better readability, the required bound is stated below as a self-contained lemma and proved in Section~\ref{pr:stein_bound}.

\begin{lemma} \label{lem:stein_bound}
Let $v_1,\ldots,v_n$ and $c_1,\ldots,c_n$ be vectors of $\R^{d_y}$ satisfying $\sum_{i\in[n]}c_i = 0$, $\max_{i,j}\|v_i-v_j\|^2\leq R$, and $\max_{i,j}\|c_i-c_j\|^2\leq R$. Then, for every $\lambda\in\R$,
    \begin{align}
  \log \E_{\pi} \exp\left(\frac{2\lambda}{n}\sum_{i=1}^n  \left\langle  v_i, c_{\pi(i)}\right\rangle \right) \leq \frac{\lambda^2 R^2}{2n},
\end{align}
where $\pi$ is uniformly chosen among all possible permutations $\{\pi\colon [n]\to [n]\}$.
\end{lemma}

Applying Lemma~\ref{lem:stein_bound} conditionally on $(S,W,\vc{U},\vc{U}')$ with $v_i=V_i$ and $c_i=C_i$, which is legitimate by \eqref{eq:C_centered} and the two diameter bounds established above, we obtain the following. Hence,
\begin{align}
    \E_{\vc{U},\vc{U}'\sim \vc{Q}} \exp\left(\frac{2\lambda}{n}\sum_{i=1}^n  \left\langle  V_i, g(U_i,W_d)-\bar{g}(\vc{U},W_d)\right\rangle \right) \leq &  \exp\left(\frac{\lambda^2 R^2}{2n}\right). \label{eq:perm_term_decomp_2_final}
\end{align}
Combining this \eqref{eq:dv_term_2_simp_1} gives
\begin{align}
    \Delta_2 \leq  & \frac{1}{\lambda} \textnormal{MDL}(\vc{Q}) + \frac{R\sqrt{2}}{\sqrt{n}} +  \frac{\lambda R^2}{2n}.   \label{eq:dv_term_2_final}
\end{align}

\textbf{Bounding the generalization error.} Plugging \eqref{eq:dv_term_1_final} and \eqref{eq:dv_term_2_final} into \eqref{eq:term_1} and \eqref{eq:term_2}, respectively, yields
\begin{align}
    \E_{S,W}\left[\gen_r(S,W)\right] \leq \frac{2}{\lambda} \textnormal{MDL}(\vc{Q}) + \frac{ \lambda R^2 }{n}+\frac{R\sqrt{2}}{\sqrt{n}}.
\end{align}	
Letting $\lambda = \sqrt{\frac{2n \textnormal{MDL}(\vc{Q}) }{R^2}}$ gives
\begin{align}
    \E_{S,W}\left[\gen_r(S,W)\right] \leq R \sqrt{\frac{8\textnormal{MDL}(\vc{Q})}{n}}  + \frac{R \sqrt{2}}{\sqrt{n}},
\end{align}	
which completes the proof.

\section{Proof of Theorem~\ref{th:distributed_gen}} \label{pr:distributed_gen}
First, following the same lines of the proof of \cite[Theorem~4]{sefidgaran2023minimum} (re-stated in Theorem~\ref{th:generalizationExp_old}), it can be seen that the prior $\vc{Q}$ is required to satisfy the symmetry property only for the permutation $\pi_{(\vc{Y},\vc{Y}')}$, defined before Theorem~\ref{th:distributed_gen}. For better readability, we recall the definition of this permutation.

Denote $\tilde{Y}^{2n}\coloneqq (\vc{Y},\vc{Y}')$ and conversely for a given $\tilde{Y}^{2n}$, let $Y_i =\tilde{Y}_i$ and  $Y'_i=\tilde{Y}_{i+n}$. We use similar notations for $\tilde{U}^{2n}$. For a given $\tilde{Y}^{2n}$, let the permutation $\pi_{\tilde{Y}^{2n}}\colon[2n]\to [2n]$, that is denoted simply as $\pi$, be the permutation with the following properties: \textbf{i.} for $i\in[n]$, $\pi(i)\in \{i\} \cup \{n+1,\ldots,2n\}$ and  $\pi(i+n)\in \{1,\ldots,n\} \cup \{i+n\}$, \textbf{ii.} $\pi(\pi(i))=i$, \textbf{iii.} $\tilde{Y}_i = 
\tilde{Y}_{\pi(i)}$, and  \textbf{iv.} it maximizes the cardinality of the set $\{i\colon \pi(i) \neq i\}$. If there exist multiple such permutations, choose one of them in a deterministic manner. 

Now, it suffices to show that
\begin{align}
    \text{MDL}(\vc{Q}) \leq  \textnormal{MDL}_{dist}(\vc{Q}_1,\ldots,\vc{Q}_K), \label{eq:pr_dist_1}
\end{align}
for some symmetric choice of the prior $\vc{Q}$, where
\begin{align}
       \textnormal{MDL}_{dist}(\vc{Q}_1,\ldots,\vc{Q}_K) \coloneqq &\sum\nolimits_{k\in[K]} \E_{S_k,S'_k,W_{e,k}} \left[ D_{KL}\left(P_{\vc{U}_k, \vc{U}'_k|\vc{X}_k,\vc{X}'_k,W_{e,k}} \Big\| \vc{Q}_k 
     \right)\right] \nonumber \\
     &- \E_{S,S',W_e}\left[D_{KL} \Big(\overline{\mathsf{P}}_{\vc{U},\vc{U}'|\vc{Y},\vc{Y}',W_e}\big\|\prod\nolimits_{k\in[K]}\vc{Q}_k\Big)\right].\label{eq:Dist_Redundancies2_dupp}
\end{align}
To show \eqref{eq:pr_dist_1}, we choose $\vc{Q}$ as $\overline{\mathsf{P}}_{\vc{U},\vc{U}'|\vc{Y},\vc{Y}',W_e}$ defined as 
\begin{align}
\overline{\mathsf{P}}_{\vc{U},\vc{U}'|\vc{Y},\vc{Y}',W_e} \coloneqq \E_{\vc{X},\vc{X}'|\vc{Y},\vc{Y}',W_e}\left[\frac{P_{\tilde{U}^{2n}|\vc{X},\vc{X}',W_e}+P_{\tilde{U}^{2n}_{\pi}|\vc{X},\vc{X}',W_e}}{2}\right].
\end{align}
It can be easily verified that $\overline{\mathsf{P}}_{\vc{U},\vc{U}'|\vc{Y},\vc{Y}',W_e}$, which is abbreviated often as $\overline{\mathsf{P}}$ for simplicity, satisfies the symmetry property with respect to $\pi_{\vc{Y},\vc{Y}'}$.

Now, we can write
\begin{align}
     \textnormal{MDL}(\overline{\mathsf{P}}) = &   \E_{S,S',W_e} \left[ D_{KL}\left(P_{\vc{U}, \vc{U}'|\vc{X},\vc{X}',W_e} \big\| \overline{\mathsf{P}} \
     \right) \right]\nonumber \\
     = & \E_{S,S',\vc{U}, \vc{U}',W_e} \left[ \log\left(\frac{P_{\vc{U}, \vc{U}'|\vc{X},\vc{X}',W_e}}{\overline{\mathsf{P}}}   
     \right) \right] \nonumber \\
     = & \E_{S,S',\vc{U}, \vc{U}',W_e} \left[ \log\left(\frac{P_{\vc{U}, \vc{U}'|\vc{X},\vc{X}',W_e}}{\overline{\mathsf{P}}}   
     \right) \right] \nonumber \\
     = & \E_{S,S',\vc{U}, \vc{U}',W_e} \left[ \log\left(\frac{P_{\vc{U}, \vc{U}'|\vc{X},\vc{X}',W_e}}{\prod\nolimits_{k\in[K]}\vc{Q}_k}   
     \right) \right]-\E_{S,S',\vc{U}, \vc{U}',W_e} \left[ \log\left(\frac{\overline{\mathsf{P}}} {\prod\nolimits_{k\in[K]}\vc{Q}_k}  
     \right) \right]  \nonumber \\
     = & \sum\nolimits_{k\in[K]} \E_{S_k,S'_k,W_{e,k}} \left[ D_{KL}\left(P_{\vc{U}_k, \vc{U}'_k|\vc{X}_k,\vc{X}'_k,W_{e,k}} \Big\| \vc{Q}_k 
     \right)\right]\nonumber\\
     &-\E_{S,S',\vc{U}, \vc{U}',W_e} \left[ \log\left(\frac{\overline{\mathsf{P}}} {\prod\nolimits_{k\in[K]}\vc{Q}_k}  
     \right) \right]. \nonumber 
\end{align}
Hence, it suffices to show that 
\begin{align}
    \E_{S,S',\vc{U}, \vc{U}',W_e} \left[ \log\left(\frac{P_{\vc{U},\vc{U}'|\vc{Y},\vc{Y}',W_e}} {\prod\nolimits_{k\in[K]}\vc{Q}_k}  
     \right) \right] = \E_{S,S',W_e}\left[D_{KL} \Big(\overline{\mathsf{P}}_{\vc{U},\vc{U}'|\vc{Y},\vc{Y}',W_e}\big\|\prod\nolimits_{k\in[K]}\vc{Q}_k\Big)\right].
\end{align}
To show this, note that the marginal priors are symmetric, and hence $\vc{Q}_k(\vc{U}_\pi,\vc{U}'_\pi|S,S',W_e)$ remains invariant under any permutation that preserves the label of $(\vc{Y},\vc{Y}')$. In particular, they are invariant under the permutation $\pi_{(\vc{Y},\vc{Y}')}$ that is defined above. As discussed above, $\overline{\mathsf{P}}$ is also invariant under such permutations. Hence,
\begin{align}
    \E_{S,S',\vc{U}, \vc{U}',W_e}  \left[ \log\left(\frac{P_{\vc{U},\vc{U}'|\vc{Y},\vc{Y}',W_e}} {\prod\nolimits_{k\in[K]}\vc{Q}_k}  
     \right) \right] = & \frac{1}{2}\E_{S,S',\vc{U}, \vc{U}',W_e}  \left[ \log\left(\frac{P_{\tilde{U}^{2n}|\vc{X},\vc{X}',W_e}} {\prod\nolimits_{k\in[K]}\vc{Q}_k}  
     \right) \right] \nonumber \\
     &+\frac{1}{2}  \E_{S,S',\vc{U}, \vc{U}',W_e}  \left[ \log\left(\frac{P_{\tilde{U}^{2n}_{\pi}|\vc{X},\vc{X}',W_e}} {\prod\nolimits_{k\in[K]}\vc{Q}_k} 
     \right) \right] \nonumber \\ 
     =& \E_{S,S'W_e} \E_{\vc{U}, \vc{U}'\sim P_{\vc{U},\vc{U}'|\vc{Y},\vc{Y}',W_e}} \left[ \log\left(\frac{P_{\vc{U},\vc{U}'|\vc{Y},\vc{Y}',W_e}}{\prod\nolimits_{k\in[K]}\vc{Q}_k} 
     \right) \right]\nonumber \\
     = &\E_{S,S',W_e}\left[D_{KL} \Big(\overline{\mathsf{P}}_{\vc{U},\vc{U}'|\vc{Y},\vc{Y}',W_e}\big\|\prod\nolimits_{k\in[K]}\vc{Q}_k\Big)\right]. \nonumber
\end{align}
This completes the proof.

\section{Proof of Lemma~\ref{lem:stein_bound}} \label{pr:stein_bound}
Lemma~\ref{lem:stein_bound} is a statement about a bilinear permutation statistic and involves no learning-theoretic object; we therefore prove it for arbitrary vectors satisfying its hypotheses. It extends \cite[Proposition~1.1]{chatterjee2006stein}, which treats the scalar statistic $\sum_i a_{i,\pi(i)}$ for a fixed array $(a_{i,j})$, to the vector-valued bilinear form $\sum_i\langle v_i,c_{\pi(i)}\rangle$ under a \emph{pairwise-diameter} constraint rather than an entry wise-boundedness constraint, which is what makes it directly applicable to decoder outputs living in a set of bounded diameter.

To prove this result, we follow a similar approach to \cite[Proposition~1.1]{chatterjee2006stein} that uses \cite[Theorem~1.5]{chatterjee2006stein}. Consider a random permutation $\pi$ and let 
\begin{align}
    \mathsf{X}  \triangleq \sum_{i=1}^n   \left\langle v_i, c_{\pi(i)}\right\rangle.
\end{align}
We use the notations $\mathsf{x},\mathsf{X},\mathsf{y},\mathsf{Y}$ only within this proof; they should not be confused with the feature and target values  $x,X,y,Y$ used elsewhere.

Note that $$\E_{\pi}[\mathsf{X}]= \sum_{i=1}^n   \left\langle v_i, \E_{\pi}\left[c_{\pi(i)}\right]\right\rangle = 0.$$ Now choose two independent random indices $(I,J) \sim \text{Uniform}(\{1,\ldots,n\})^{\otimes 2}$ and let $\pi'$ be the permutation obtained by transposition of $\pi(I)$ and $\pi(J)$, i.e. i) $\pi'(r) = \pi(r)$ for $r\notin \{I,J\}$, ii) $\pi'(I)=\pi(J)$, and iii) $\pi'(J)=\pi(I)$. Let 
\begin{align}
    \mathsf{X}' \triangleq\sum_{i=1}^n  \left\langle  v_i, c_{\pi'(i)}\right\rangle.
\end{align}
Since $\pi$ is uniform and $\pi'$ is obtained via a random transposition, $(\pi,\pi')$ is exchangeable, and hence so is $(\mathsf{X}, \mathsf{X}')$.

Let $f(\mathsf{x}) = \mathsf{x}$ and $F(\mathsf{x},\mathsf{y}) = \frac{n}{2}(\mathsf{x}-\mathsf{y})$. Then, 
\begin{align}
    \mathbb{E}\left[ F(\mathsf{X},\mathsf{X}') \mid \pi \right] =& \frac{n}{2} \E_{I,J}\left[\left\langle v_I-v_J,  c_{\pi(I)}- c_{\pi(J)} \right\rangle  \mid \pi \right] \nonumber \\
    \stackrel{(a)}{=}&  \frac{n}{2}\E_{I}\left[\left\langle v_I,  c_{\pi(I)} \right\rangle  \mid \pi\right]+ \frac{n}{2}\E_{J}\left[\left\langle v_J,  c_{\pi(J)} \right\rangle  \mid \pi\right]  \nonumber \\
    =&  \frac{1}{2}\sum_{i=1}^n\left\langle v_i,  c_{\pi(i)} \right\rangle +  \frac{1}{2}\sum_{j=1}^n\left\langle v_j,  c_{\pi(j)} \right\rangle \nonumber \\
    =& f(\mathsf{X}),  \label{eq:steinCondition}
\end{align}
where $(a)$ uses the fact that $\sum_{i\in[n]}c_i =0$, which implies $\sum_{j\in[n]}c_{\pi(j)} =0$.

Since $\mathsf{X}$ is a measurable function of $\pi$, the tower property yields
\begin{align}
\mathbb{E}[F(\mathsf{X}, \mathsf{X}') \mid \mathsf{X}] = \mathbb{E}[ \mathbb{E}[F(\mathsf{X}, \mathsf{X}') \mid \pi] \mid \mathsf{X} ] = \mathbb{E}[f(\mathsf{X}) \mid \mathsf{X}] = f(\mathsf{X}).    \label{eq:tower_chain} 
\end{align}
Now, since \eqref{eq:steinCondition} holds and since $F(\mathsf{x},\mathsf{y})$ is antisymmetric (i.e. $F(\mathsf{x},\mathsf{y})=-F(\mathsf{y},\mathsf{x})$), we can use \cite[Theorem~1.5]{chatterjee2006stein}. Let
\begin{align}
    \Delta(\mathsf{X}) \triangleq \frac{1}{2} \E\left[|(f(\mathsf{X}) - f(\mathsf{X}')) F(\mathsf{X},\mathsf{X}')| \mid \mathsf{X} \right].
\end{align}
Then, first note that
\begin{align}
   \frac{1}{2} \E\left[|(f(\mathsf{X}) - f(\mathsf{X}')) F(\mathsf{X},\mathsf{X}')| \mid \pi \right]  = & \frac{n}{4} \E\left[(\mathsf{X}-\mathsf{X}')^2 \mid\pi\right] \nonumber \\
    = & \frac{1}{4n} \sum_{i=1}^n \sum_{j=1}^n  \left\langle v_i-v_j, c_{\pi(i)}-c_{\pi(j)} \right\rangle^2 \nonumber \\
    \stackrel{(a)}{\leq} & \frac{1}{4n} \sum_{i=1}^n \sum_{j=1}^n  \|v_i-v_j\|^2 \|c_{\pi(i)}-c_{\pi(j)}\|^2 \nonumber \\    
    \stackrel{(b)}{\leq} & \frac{nR^2}{4},
\end{align}
where $(a)$ follows by the Cauchy-Schwarz inequality and $(b)$ by the hypotheses $\|c_{\pi(i)}-c_{\pi(j)}\|^2 \leq R$ and $\|v_i-v_j\|^2\leq R$ of the lemma.

Now, again similar to \eqref{eq:tower_chain}, we have
\begin{align}
    \Delta(\mathsf{X}) = \mathbb{E}\left[\frac{1}{2} \mathbb{E}\left[|(f(\mathsf{X}) - f(\mathsf{X}')) F(\mathsf{X},\mathsf{X}')| \mid \pi\right] \mid \mathsf{X} \right]   \leq &  \frac{nR^2}{4} \label{eq:steinDelta}
\end{align}

Since \eqref{eq:steinDelta} is a constant for every $\pi$, we have $\|\Delta(\mathsf{X})\|_{\infty}\leq \frac{nR^2}{4}\eqqcolon \kappa$. Finally, using the bound on the MGF, developed in the proof of \cite[Theorem~1.5]{chatterjee2006stein}, we have
\begin{align}
  \log \E_{\pi} \exp\left(\frac{2\lambda}{n}\sum_{i=1}^n  \left\langle  v_i, c_{\pi(i)}\right\rangle \right) \leq  \frac{1}{2}\|\Delta(\mathsf{X})\|_{\infty}  \left(\frac{2\lambda}{n}\right)^2  \leq  \frac{1}{2}\kappa \left(\frac{2\lambda}{n}\right)^2 = \frac{\lambda^2 R^2}{2n}.
\end{align}
This completes the proof.

\bibliographystyle{IEEEtran}
\bibliography{IEEEabrv,biblio}

\vspace{0.5 cm}
\begin{center}
    {\huge \textbf{Appendices}}
\end{center}
The appendices are organized as follows:

\begin{itemize}
    \item In Appendix~\ref{sec:redundancyExample} we show by an example how the joint regularizer, introduced in Section~\ref{sec:joint_multivew_regularizer}, penalizes cross-view redundancies in the latent variables less than the marginal-only regularizer of Section~\ref{sec:marginal_reg}.
    \item Appendix~\ref{sec:single_view} provides a self-contained summary of the single-view Gaussian mixture approach of \cite{sefidgaran2025generalization}, included here for completeness.
    \item In Appendix~\ref{sec:average_estimate}, we explain in detail our approach to finding the Gaussian product mixture priors and how to use them in a regularizer term. This section is divided into three parts, describing our initialization method, followed by the lossless and lossy versions of our approach.
         \item  Appendix~\ref{sec:details_exp} contains further details about the experiments reported in this paper, as well as 15 supplementary experiments.

    \end{itemize}

\appendices



\section{On penalizing redundancies in multiview regularizers} \label{sec:redundancyExample}

The joint regularizer proposed in Section~\ref{sec:joint_multivew_regularizer} considers the MDL of all views simultaneously. Here, we verify that it penalizes cross-view redundancies in the latent variables less than the marginal-only regularizer of Section~\ref{sec:marginal_reg}, which treats each view independently.

To illustrate this, consider $K=2$ and suppose that for every input $x=(x_1,x_2)$, the latent variable parameters satisfy $\sigma_{x,1}=\sigma_{x,2}\eqqcolon\sigma_x$ and $\mu_{x,1}=\mu_{x,2}\eqqcolon\mu_x$. Note that while the parameters are identical, the latent variables are not: both are independent draws from $\mathcal{N}(\mu_x,\diag(\sigma_x^2))$, so they are statistically redundant but not identical. Suppose further that for every $c\in[C]$, the pair $(\mu_x,\sigma_x)$ equals $(\mu_{c,r},\sigma_{c,r})$ with probability $\beta_{c,r}\in[0,1]$ for $r\in[R]$, with $\sum_{r\in[R]}\beta_{c,r}=1$. Setting $M=R$ and neglecting the symmetry condition for simplicity, the optimal marginal priors are
\begin{align*}
Q_{c,1}=Q_{c,2}=\sum_{r\in[R]}\beta_{c,r}\,\mathcal{N}\big(\mu_{c,r},\diag(\sigma_{c,r}^2)\big),
\end{align*}
and the optimal joint prior is
\begin{align*}
Q_c=\sum_{r\in[R]}\beta_{c,r}\,\mathcal{N}\big(\mu_{c,r},\diag(\sigma_{c,r}^2)\big)\mathcal{N}\big(\mu_{c,r},\diag(\sigma_{c,r}^2)\big).
\end{align*}
In the optimal joint prior, the marginal components always share the same mean and covariance. Denoting the joint regularizer~\eqref{eq:reg_lossless_multi} as $R_2$ and the marginal-only regularizer as $R_1$, a comparison for $b=\infty$ gives
\begin{align}
    R_2 =& - \sum_{c\in[C]} \sum_{r\in[R]} P(Y=c) \beta_{c,r} \log\left(\sum_{r'\in[R]} \beta_{c,r'} e^{-2 D_{KL}\Big(\mathcal{N}\big(\mu_{c,r},\diag(\sigma_{c,r}^2)\big)\big\|\mathcal{N}\big(\mu_{c,r'},\diag(\sigma_{c,r'}^2)\big)\Big)} \right)\nonumber \\
    \stackrel{(a)}{\leq}& {-} \sum_{c\in[C]} \sum_{r\in[R]}P(Y{=}c) \beta_{c,r} \log\left( \left(\sum_{r'\in[R]}  \beta_{c,r'} e^{- D_{KL}\Big(\mathcal{N}\big(\mu_{c,r},\diag(\sigma_{c,r}^2)\big)\big\|\mathcal{N}\big(\mu_{c,r'},\diag(\sigma_{c,r'}^2)\big)\Big)} \right)^2 \right) \nonumber \\
    =& -2 \sum_{c\in[C]} \sum_{r\in[R]} P(Y=c) \beta_{c,r} \log \left(\sum_{r'\in[R]} \beta_{c,r'} e^{-D_{KL}\Big(\mathcal{N}\big(\mu_{c,r},\diag(\sigma_{c,r}^2)\big)\big\|\mathcal{N}\big(\mu_{c,r'},\diag(\sigma_{c,r'}^2)\big)\Big)}  \right)\nonumber\\
    = & \, R_1,
\end{align}
where $(a)$ follows from the convexity of $f(x)=x^2$ and Jensen's inequality $-\log\mathbb{E}[f(X)]\leq-\log f(\mathbb{E}[X])$. Hence, $R_2\leq R_1$: the joint regularizer~\eqref{eq:reg_lossless_multi} penalizes cross-view redundancies strictly less than the marginal-only regularizer, confirming it as the more appropriate choice.

\section{Single-view Gaussian mixture prior: a summary of \texorpdfstring{\cite{sefidgaran2025generalization}}{[Sefidgaran et al., 2025]}}
\label{sec:single_view}

The approach of \cite{sefidgaran2025generalization} simultaneously finds and imposes a data-dependent Gaussian mixture prior for the single-view setup ($K=1$). For $c\in[C]$, let $Q_c$ denote the data-dependent Gaussian mixture prior for target $c$, and let $\vc{Q}(\vc{U},\vc{U'}|S,S',\hat{W}_e)=\prod_{i\in[n]} Q_{Y_i}(U_i)Q_{Y'_i}(U'_i)$, which satisfies the conditional symmetry property of Definition~\ref{def:symmetry}. The approach applies to both lossless and lossy bounds; the lossy approach is reported to provide better performance often in the experiments. For brevity, we provide only a summary of the lossless approach and refer the reader to \cite{sefidgaran2025generalization} for the lossy approach.
\subsection{Lossless Gaussian mixture prior}
\label{app:lossless_brief}

For every $c\in[C]$, the prior $Q_c$ is defined over $\mathbb{R}^d$ as
\begin{equation}
   Q_{c} = \sum\nolimits_{m\in[M]} \alpha_{c,m}\, Q_{c,m}, \quad Q_{c,m}=\mathcal{N}\left(\mu_{c,m},\diag\left(\sigma_{c,m}^2\right)\right),\quad m\in[M],\; c\in [C],
   \label{eq:prior-model-single-view}
\end{equation}
where $\{\alpha_{c,m}\}_{m}$ are non-negative with $\sum_{m\in[M]} \alpha_{c,m}=1$. Since the KL-divergence between a Gaussian and a Gaussian mixture has no closed form, it is estimated as $\big(D_{\text{var}}+D_{\text{prod}}\big)/2$, where $D_{\text{var}}$ and $D_{\text{prod}}$ are respectively the variational lower bound and the product Gaussians upper bound of \cite{hershey2007approximating}; see \cite[Appendices~C,~F]{sefidgaran2025generalization} for details. For clarity of exposition, only $D_{\text{var}}$ is used below.

Considering only the training part of $\textnormal{MDL}(\vc{Q})$ and restricting to a mini-batch $\mathcal{B}=\{z_1,\ldots,z_b\}\subseteq S$, the regularizer is
\begin{align}
    \textnormal{Regularizer}(\vc{Q}) \coloneqq  D_{KL}\big(P_{\vc{U}_{\mathcal{B}}|\vc{X}_{\mathcal{B}},W_e} \big\| \vc{Q}_{\mathcal{B}}\big). \label{eq:regularizer}
\end{align}
The prior components are initialized similarly to k-means++ \cite{arthur2007k}; see \cite[Appendix~C.1]{sefidgaran2025generalization} for details. At iteration $(t)$, the regularizer admits the upper bound
\begin{align}
    \textnormal{Regularizer}(\vc{Q}) \leq \sum\limits_{i\in[b]} \sum\limits_{m\in[M]} \gamma_{i,m} \left(D_{KL}\big(P_{U_i|x_i,w_e} \big\| Q^{(t)}_{y_i,m}(U_i)\big)-\log\big(\alpha_{y_i,m}^{(t)}/\gamma_{i,m}\big)\right),  \label{eq:reg_lossless_single_not_simplified}
\end{align}
for any $\gamma_{i,m}\geq 0$ with $\sum_{m\in[M]} \gamma_{i,m} =1$; see \cite{sefidgaran2025generalization}. The minimizing coefficients are
\begin{align}
    \gamma_{i,m} = \frac{\alpha_{y_i,m}^{(t)} e^{-D_{KL}\big(P_{U_i|x_i,w_e}\|Q_{y_i,m}^{(t)}\big)}}{\sum\nolimits_{m'\in [M]} \alpha_{y_i,m'}^{(t)} e^{-D_{KL}\big(P_{U_i|x_i,w_e}\|Q_{y_i,m'}^{(t)}\big)}}, \quad i\in[b],\; m\in[M], \label{eq:gamma_i}
\end{align}
and setting $\gamma_{i,c,m} = \gamma_{i,m}$ if $c=y_i$ and $0$ otherwise, the optimal parameters are
\begin{align}
    \mu_{c,m}^* =& \frac{1}{b_{c,m}}\sum\nolimits_{i\in[b]}  \gamma_{i,c,m} \mu_{x_i},  \quad {\sigma^{*}_{c,m,j}}^2 = \frac{1}{b_{c,m}}\sum\nolimits_{i\in[b]}  \gamma_{i,c,m} \left(\sigma_{x_i,j}^2+(\mu_{x_i,j}-\mu_{c,m,j}^{(t)})^2\right),\nonumber\\
    \alpha_{c,m}^* = & b_{c,m}/b_c, \quad b_{c,m}=\sum\nolimits_{i\in[b]}  \gamma_{i,c,m}, \quad b_c =\sum\nolimits_{m\in[M]} b_{c,m}.\label{eq:optimal_updates_lossless_single}
\end{align}
The prior parameters are then updated via moving averages
\begin{align}
    \mu_{c,m}^{(t+1)} =& (1-\eta_1) \mu_{c,m}^{(t)}+\eta_1 \mu_{c,m}^*+\mathfrak{Z}_{1}^{(t+1)}, \quad {\sigma_{c,m}^{(t+1)}}^2 = (1-\eta_2) {\sigma_{c,m}^{(t)}}^2+\eta_2 {\sigma_{c,m}^*}^2+\mathfrak{Z}_{2}^{(t+1)},\nonumber \\
    \alpha_{c,m}^{(t+1)} = &(1-\eta_3) \alpha_{c,m}^{(t)}+\eta_3 \alpha_{c,m}^*, \label{eq:updates_lossless_single}
\end{align}
where $\eta_1,\eta_2,\eta_3 \in [0,1]$ and $\mathfrak{Z}_{j}^{(t+1)}\sim\mathcal{N}(\vc{0}_d,\zeta_j^{(t+1)}\mathrm{I}_d)$, with $\zeta_j^{(t+1)}\in \mathbb{R}_+$ fixed. Using \eqref{eq:gamma_i}, the regularizer simplifies to
\begin{equation}
    - \sum\nolimits_{i\in[b]} \log\Big(\sum\nolimits_{m\in [M]} \alpha_{y_i,m}^{(t)} e^{-D_{KL}\big(P_{U_i|x_i,w_e}\|Q_{y_i,m}^{(t)}\big)} \Big).\label{eq:reg_lossless_single}
\end{equation}


\section{Gaussian product mixture prior: approximation and regularization} \label{sec:average_estimate}

This section details the approach to constructing a data-dependent Gaussian product mixture prior and deploying it as a regularizer along the optimization trajectory. For the single-view approach, the reader is referred to \cite[Appendix~C]{sefidgaran2025generalization} and to Appendix~\ref{sec:single_view} for a summary. The section is organized as follows: Appendix~\ref{sec:initialization_multi_view} describes the initialization, and Appendices~\ref{sec:lossless_multi_view} and~\ref{sec:lossy_multi_view} cover the lossless and lossy versions, respectively.

The regularizer considered throughout is
\begin{align}
    \textnormal{Regularizer}(\vc{Q}) \coloneqq D_{KL}\big(P_{\vc{U}_{\mathcal{B}}|\vc{X}_{\mathcal{B}},W_e}\big\|\vc{Q}_{\mathcal{B}}\big), \label{eq:regularizer_multi}
\end{align}
where $\mathcal{B}$ denotes the current mini-batch and the dependence on $\mathcal{B}$ is dropped for notational convenience. The superscript $(t)$ denotes iteration $t\in\mathbb{N}^*$. Recall that $\vc{X}=(\vc{X}_1,\ldots,\vc{X}_K)$, $\vc{U}=(\vc{U}_1,\ldots,\vc{U}_K)$, and $W_e=(W_{e,1},\ldots,W_{e,K})$.

In both the lossless and lossy approaches, three sets of parameters $\alpha_{c,\vc{m}}^{(0)}$, $\mu_{c,k,m}^{(0)}$, and $\sigma_{c,k,m}^{(0)}$ are initialized for $c\in[C]$, $\vc{m}=(m_1,\ldots,m_K)\in[M]^K$, $k\in[K]$, and $m\in[M]$, as described next.

\subsection{Initialization of the components} \label{sec:initialization_multi_view}

The joint coefficients are initialized as $\alpha_{c,\vc{m}}^{(0)}=M^{-K}$ for all $c\in[C]$ and $\vc{m}\in[M]^K$. The standard deviations $\sigma_{c,k,m}^{(0)}$ are drawn independently from $\mathcal{N}(0,\mathrm{I}_d)$.

The component means $\{\mu_{c,k,m}^{(0)}\}_{k\in[K]}$ are initialized jointly across all views via a distributed adaptation of the k-means$++$ seeding strategy \cite{arthur2007k}, as follows.

\begin{itemize}
    \item[1.] The encoders $W_e=(W_{e,1},\ldots,W_{e,K})$ are initialized independently.
    \item[2.] A large mini-batch $\vc{Z}=\{Z_1,\ldots,Z_{\tilde{b}}\}$ with $\tilde{b}\gg b$ is drawn from the training data. Let $\vc{X}$ and $\vc{Y}$ denote the corresponding features and targets, with $\vc{X}=(\vc{X}_1,\ldots,\vc{X}_K)$. For each $c\in[C]$, let $\vc{X}_c=\{X_{c,1},\ldots,X_{c,b_c}\}\subseteq\vc{X}$ be the subset with target $c$, where $\sum_{c\in[C]}b_c=\tilde{b}$ and $X_{c,i}=(X_{c,1,i},\ldots,X_{c,K,i})$.

    Using the initialized encoders, compute the latent mean vectors for this mini-batch. Denote them as $\boldsymbol{\mu}_c=\{\mu_{c,1},\ldots,\mu_{c,b_c}\}$ with $\mu_{c,i}=(\mu_{c,1,i},\ldots,\mu_{c,K,i})$. For each $c\in[C]$, the first component means across all views $\left(\mu_{c,1,1}^{(0)},\ldots,\mu_{c,K,1}^{(0)}\right)$ are set to a uniformly sampled element of $\boldsymbol{\mu}_c$, so that all first components correspond to the latent parameters of a single training sample across views.

    \item[3.] For $2\leq m\leq M$, draw a new mini-batch with per-target features $\vc{X}_c$ and latent means $\boldsymbol{\mu}_c$. For each $c\in[C]$, compute
    \begin{align*}
        d_{\min,c}(i) = \sum_{k\in[K]}\min_{m'_k\in[m-1]}\left\|\mu_{c,k,i}-\mu_{c,k,m'}^{(0)}\right\|^2, \quad i\in[b_c].
    \end{align*}
    Sample $i^*$ from $[b_c]$ with probability proportional to $d_{\min,c}(i)$, and set $\left(\mu_{c,1,m}^{(0)},\ldots,\mu_{c,K,m}^{(0)}\right)=\mu_{c,i^*}$. As before, all $m$-th component means correspond to the latent parameters of a single sample across views.
\end{itemize}

\subsection{Lossless Gaussian product mixture} \label{sec:lossless_multi_view}

For each $c\in[C]$, the prior $Q_c$ over $\mathbb{R}^{Kd}$ is defined as
\begin{align}
   Q_c = \sum_{\vc{m}\in[M]^K}\alpha_{c,\vc{m}}\,Q_{c,\vc{m}},
\end{align}
where $\vc{m}=(m_1,\ldots,m_K)$, $\alpha_{c,\vc{m}}\in[0,1]$, $\sum_{\vc{m}\in[M]^K}\alpha_{c,\vc{m}}=1$, and
\begin{align*}
Q_{c,\vc{m}}=\prod_{k\in[K]}Q_{c,k,m_k}, \qquad Q_{c,k,m}=\mathcal{N}\left(\mu_{c,k,m},\diag\left(\sigma_{c,k,m}^2\right)\right), \quad k\in[K],\,m\in[M],\,c\in[C].
\end{align*}
Defining $\alpha_{c,k,m}=\sum_{\vc{m}\in[M]^K\colon m_k=m}\alpha_{c,\vc{m}}$, the marginal prior of view $k$ under $Q_c$ is
\begin{align*}
    Q_{c,k}=\sum_{m\in[M]}\alpha_{c,k,m}\,Q_{c,k,m},
\end{align*}
which is a Gaussian mixture, consistent with \cite{sefidgaran2025generalization}.

\textbf{Update of the priors.} At iteration $(t)$ with mini-batch $\mathcal{B}^{(t)}=\{z_1^{(t)},\ldots,z_b^{(t)}\}$ (dependence on $(t)$ dropped), the regularizer equals $\textnormal{Regularizer}(\vc{Q})=\sum_{i\in[b]}D_{KL}\big(P_{U_i|x_i,w_e}\big\|Q^{(t)}_{y_i}\big)$. Upper and lower bounds are derived as follows.

The variational upper bound is
\begin{align}
    &\textnormal{Regularizer}(\vc{Q})\leq D_{\text{var}} \coloneqq\nonumber \\
    & \sum_{i\in[b]}\sum_{\vc{m}\in[M]^K}\gamma_{i,\vc{m}}\left(\sum_{k\in[K]}D_{KL}\big(P_{U_{i,k}|x_{i,k},w_{e,k}}\big\|Q^{(t)}_{y_i,k,m_k}(U_{i,k})\big)-\log\big(\alpha_{y_i,\vc{m}}^{(t)}/\gamma_{i,\vc{m}}\big)\right), \label{eq:reg_lossless_multi_var}
\end{align}
for any $\gamma_{i,\vc{m}}\geq 0$ with $\sum_{\vc{m}\in[M]^K}\gamma_{i,\vc{m}}=1$ for each $i\in[b]$. The minimizing coefficients are
\begin{align*}
    \gamma_{i,\vc{m}}=\frac{\alpha_{y_i,\vc{m}}^{(t)}e^{-\sum_{k\in[K]}D_{KL}\big(P_{U_{i,k}|x_{i,k},w_{e,k}}\big\|Q^{(t)}_{y_i,k,m_k}(U_{i,k})\big)}}{\sum_{\vc{m}'\in[M]^K}\alpha_{y_i,\vc{m}'}^{(t)}e^{-\sum_{k\in[K]}D_{KL}\big(P_{U_{i,k}|x_{i,k},w_{e,k}}\big\|Q^{(t)}_{y_i,k,m'_k}(U_{i,k})\big)}}, \quad i\in[b],\,\vc{m}\in[M]^K.
\end{align*}
Setting $\gamma_{i,c,\vc{m}}=\gamma_{i,\vc{m}}$ if $c=y_i$ and $0$ otherwise, the estimated lower bound is
\begin{align}
    \textnormal{Regularizer}&(\vc{Q})\geq D_{\text{prod}}\coloneqq\nonumber \\
    &-\sum_{i\in[b]}\left(\frac{1}{2}\log\Big((2\pi e)^{Kd}\prod_{k\in[K]}\prod_{j\in[d]}\sigma_{x_{i,k},j}^2\Big)+\log\Big(\sum_{\vc{m}\in[M]^K}\alpha_{y_i,\vc{m}}^{(t)}t'_{i,\vc{m}}\Big)\right), \label{eq:reg_lossless_multi_prod}
\end{align}
where
\begin{align}
    t'_{i,\vc{m}}\coloneqq\frac{e^{-\sum_{k\in[K]}\sum_{j\in[d]}\frac{\left(\mu_{x_i,k,j}-\mu_{y_i,k,m_k,j}^{(t)}\right)^2}{2{\sigma_{y_i,k,m_k,j}^{(t)}}^2}}}{\sqrt{\prod_{k\in[K]}\prod_{j\in[d]}\left(2\pi{\sigma_{y_i,k,m_k,j}^{(t)}}^2\right)}}. \label{eq:Gaussian_integral_multi}
\end{align}
Following \cite{hershey2007approximating,durrieu2012lower}, a more accurate estimate is obtained as
\begin{align}
    \textnormal{Regularizer}(\vc{Q})\approx\frac{D_{\text{var}}+D_{\text{prod}}}{2}\eqqcolon D_{\text{est}}. \label{eq:reg_lossless_multi_estimate} 
\end{align}
Treating $\gamma_{i,\vc{m}}$ as fixed, the parameters minimizing $D_{\text{est}}$ subject to $\sum_{\vc{m}}\alpha_{c,\vc{m}}=1$ are found by setting partial derivatives to zero. Introducing
\begin{align}
    \beta_{i,c,\vc{m}}&=\begin{cases}\frac{\eta_{i,\vc{m}}}{\sum_{\vc{m}'\in[M]^K}\eta_{i,\vc{m}'}},&\text{if }c=y_i,\\0,&\text{otherwise},\end{cases} \qquad \eta_{i,\vc{m}}\coloneqq\alpha_{y_i,\vc{m}}^{(t)}e^{-\sum_{k\in[K]}\sum_{j\in[d]}\frac{\left(\mu_{x_i,k,j}-\mu_{y_i,k,m_k,j}^{(t)}\right)^2}{2{\sigma_{y_i,k,m_k,j}^{(t)}}^2}},
\end{align}
and the marginals $\gamma_{i,c,k,m}=\sum_{\vc{m}\in[M]^K\colon m_k=m}\gamma_{i,c,\vc{m}}$, $\beta_{i,c,k,m}=\sum_{\vc{m}\in[M]^K\colon m_k=m}\beta_{i,c,\vc{m}}$, $\tilde{\gamma}_{i,c,\vc{m}}=(\gamma_{i,c,\vc{m}}+\beta_{i,c,\vc{m}})/2$, $\tilde{\gamma}_{i,c,k,m}=(\gamma_{i,c,k,m}+\beta_{i,c,k,m})/2$, the optimal parameters are
\begin{align}
    \mu_{c,k,m}^*&=\frac{1}{\tilde{b}_{c,k,m}}\sum_{i\in[b]}\tilde{\gamma}_{i,c,k,m}\,\mu_{x_i,k}, \nonumber \\
    {\sigma_{c,k,m,j}^*}^2&=\frac{1}{b_{c,k,m}}\sum_{i\in[b]}\left(\gamma_{i,c,k,m}\,\sigma_{x_i,k,j}^2+2\tilde{\gamma}_{i,c,k,m}(\mu_{x_i,k,j}-\mu_{c,k,m,j}^{(t)})^2\right),\nonumber\\
\alpha_{c,\vc{m}}^*&=\tilde{b}_{c,\vc{m}}/\tilde{b}_c,  \nonumber \\
\tilde{b}_{c,\vc{m}}&=\sum_{i\in[b]}\tilde{\gamma}_{i,c,\vc{m}}, \nonumber \\
\tilde{b}_c&=\sum_{\vc{m}\in[M]^K}\tilde{b}_{c,\vc{m}},  \nonumber \\
\tilde{b}_{c,k,m}&=\sum_{i\in[b]}\tilde{\gamma}_{i,c,k,m},  \nonumber \\
b_{c,k,m}&=\sum_{i\in[b]}\gamma_{i,c,k,m}. \label{eq:optimal_updates_lossless_multi}
\end{align}
Crucially, $\mu_{c,k,m}^*$ and $\sigma_{c,k,m}^*$ depend only on the marginal coefficients $\gamma_{i,c,k,m}$, $\beta_{i,c,k,m}$, and the $k$-th view's latent parameters. The parameters are updated via moving averages:
\begin{align}
    \mu_{c,k,m}^{(t+1)}&=(1-\eta_1)\mu_{c,k,m}^{(t)}+\eta_1\mu_{c,k,m}^*+\mathfrak{Z}_1^{(t+1)}, \quad {\sigma_{c,k,m}^{(t+1)}}^2=(1-\eta_2){\sigma_{c,k,m}^{(t)}}^2+\eta_2{\sigma_{c,k,m}^*}^2+\mathfrak{Z}_2^{(t+1)},\nonumber\\
    \alpha_{c,\vc{m}}^{(t+1)}&=(1-\eta_3)\alpha_{c,\vc{m}}^{(t)}+\eta_3\alpha_{c,\vc{m}}^*, \label{eq:updates_lossless_multi_app}
\end{align}
where $\eta_1,\eta_2,\eta_3\in[0,1]$ and $\mathfrak{Z}_j^{(t+1)}\sim\mathcal{N}(\vc{0}_d,\zeta_j^{(t+1)}\mathrm{I}_d)$ with $\zeta_j^{(t+1)}\in\mathbb{R}_+$ fixed.

\textbf{Distributed protocol.} Since $\mu_{c,k,m}^*$ and $\sigma_{c,k,m}^*$ depend only on marginal quantities, the server only needs to coordinate the joint coefficients $\gamma_{i,c,\vc{m}}$ and $\beta_{i,c,\vc{m}}$, send their marginals back to each client, and each client can update its local prior independently:
\begin{itemize}[leftmargin=*]
    \item Each client $k$ shares $D_{KL}\big(P_{U_{i,k}|x_{i,k},w_{e,k}}\big\|Q^{(t)}_{y_i,k,m_k}(U_{i,k})\big)$ for $i\in[b]$ and $m\in[M]$.
    \item The server updates $\alpha_{c,\vc{m}}^{(t+1)}$, sends the marginals $\alpha_{c,k,m}^{(t+1)}$ to each client $k$, computes the regularization term, performs backpropagation, and returns gradient vectors to the clients.
    \item Each client updates its local marginal prior and encoder using the received gradients and marginal coefficients $\alpha_{c,k,m}^{(t+1)}$.
\end{itemize}

\textbf{Regularizer.} The estimate~\eqref{eq:reg_lossless_multi_estimate} simplifies to
\begin{align}
    \textnormal{Regularizer}(\vc{Q})=&-\frac{1}{2}\sum_{i\in[b]}\log\left(\sum_{\vc{m}\in[M]^K}\alpha_{y_i,\vc{m}}^{(t)}e^{-\sum_{k\in[K]}D_{KL}\big(P_{U_{i,k}|x_{i,k},w_{e,k}}\|Q_{y_i,k,m_k}^{(t)}\big)}\right)\nonumber\\
    &-\frac{1}{2}\sum_{i\in[b]}\left(\frac{1}{2}\log\Big((2\pi e)^{Kd}\prod_{k\in[K]}\prod_{j\in[d]}\sigma_{x_{i,k},j}^2\Big)+\log\Big(\sum_{\vc{m}\in[M]^K}\alpha_{y_i,\vc{m}}^{(t)}t'_{i,\vc{m}}\Big)\right).
\end{align}

\subsection{Lossy Gaussian product mixture} \label{sec:lossy_multi_view}

The lossy regularizer accounts for the MDL of perturbed latent variables while passing unperturbed ones to the decoder. For $k\in[K]$, fix $\epsilon_k\in\mathbb{R}_+$ and let $\boldsymbol{\epsilon}_k=(\epsilon_k,\ldots,\epsilon_k)\in\mathbb{R}^d$. The perturbed variable is defined as
\begin{align}
    \hat{U}_k=(\mu_{X,k}+Z_{k,1})+Z_{k,2}\eqqcolon\hat{U}_{k,1}+\hat{U}_{k,2},
\end{align}
where $Z_{k,1}\sim\mathcal{N}\left(\vc{0}_d,\sqrt{Kd/4}\,\mathrm{I}_d\right)$ and $Z_{k,2}\sim\mathcal{N}\left(\vc{0}_d,\diag\big(\sigma_{X,k}^2+\boldsymbol{\epsilon}_k\big)\right)$ are independent, yielding $\hat{U}_{k,1}\sim\mathcal{N}(\mu_{X,k},\sqrt{Kd/4}\,\mathrm{I}_d)$ and $\hat{U}_{k,2}\sim\mathcal{N}(\vc{0}_d,\diag(\sigma_{X,k}^2+\boldsymbol{\epsilon}_k))$, conditionally independent given $(X,W_e)$. Let $\hat{U}_1=(\hat{U}_{1,1},\ldots,\hat{U}_{K,1})$ and $\hat{U}_2=(\hat{U}_{1,2},\ldots,\hat{U}_{K,2})$.

For each $c\in[C]$, two Gaussian product mixture priors $Q_{c,1}$ and $Q_{c,2}$ are defined for $\hat{U}_1$ and $\hat{U}_2$:
\begin{align}
    Q_{c,1}=\sum_{\vc{m}\in[M]^K}\alpha_{c,\vc{m}}\,Q_{c,\vc{m},1}, \qquad Q_{c,2}=\sum_{\vc{m}\in[M]^K}\alpha_{c,\vc{m}}\,Q_{c,\vc{m},2},
\end{align}
with $Q_{c,\vc{m},l}=\prod_{k\in[K]}Q_{c,k,m_k,l}$ and marginal components
\begin{align*}
Q_{c,k,m,1}=\mathcal{N}\left(\mu_{c,k,m},\sqrt{Kd/4}\,\mathrm{I}_d\right), \qquad Q_{c,k,m,2}=\mathcal{N}\left(\vc{0}_d,\diag\left(\sigma_{c,k,m}^2+\boldsymbol{\epsilon}_k\right)\right).
\end{align*}
Both priors share the same coefficients $\alpha_{c,\vc{m}}$. The prior $Q_c$ is defined as the distribution of $\hat{U}=\hat{U}_1+\hat{U}_2$ when $\hat{U}_1\sim Q_{c,1}$ and $\hat{U}_2\sim Q_{c,2}$. The lossy KL divergence is defined as
\begin{align}    D_{KL,\text{Lossy}}\left(P_{\hat{U}|x,w_e}\|Q_{y_i}\right)\coloneqq & D_{KL}\left(\mathcal{N}(\mu_x,\sqrt{Kd/4}\,\mathrm{I}_d)\|Q_{y_i,1}\right)\nonumber \\
&+D_{KL}\left(\mathcal{N}(\vc{0}_d,\diag(\sigma_x^2+\boldsymbol{\epsilon}_k))\|Q_{y_i,2}\right).
\end{align}
Using the inequality $D_{KL}(P_{\hat{U}|x,w_e}\|Q_{y_i})\leq D_{KL,\text{Lossy}}(P_{\hat{U}|x,w_e}\|Q_{y_i})$, the variational upper bound becomes
\begin{align}
    \textnormal{Regularizer}(\vc{Q})\leq D_{\text{var}}\coloneqq\sum_{i\in[b]}\sum_{\vc{m}\in[M]^K}\gamma_{i,\vc{m}}\left(D_{KL,\text{lossy}}\big(P_{U_i|x_i,w_e}\big\|Q^{(t)}_{y_i,\vc{m}}(U_i)\big)-\log\frac{\alpha_{y_i,\vc{m}}^{(t)}}{\gamma_{i,\vc{m}}}\right), \label{eq:reg_lossy_multi_var}
\end{align}
minimized at
\begin{align}
    \gamma_{i,\vc{m}}=\frac{\alpha_{y_i,\vc{m}}^{(t)}e^{-\sum_{k\in[K]}D_{KL,\text{Lossy}}\big(P_{U_i|x_i,w_e}\|Q_{y_i,\vc{m}}^{(t)}\big)}}{\sum_{\vc{m}'\in[M]^K}\alpha_{y_i,\vc{m}'}^{(t)}e^{-\sum_{k\in[K]}D_{KL,\text{Lossy}}\big(P_{U_i|x_i,w_e}\|Q_{y_i,\vc{m}'}^{(t)}\big)}}.
\end{align}
The estimated lower bound is
\begin{align}
    D_{\text{prod}}\coloneqq-\sum_{i\in[b]}\left(Kd\log\left(\pi e\sqrt{Kd}\right)+\log\Big(\sum_{\vc{m}\in[M]^K}\alpha_{y_i,\vc{m}}^{(t)}\tilde{t}_{i,\vc{m}}\Big)\right), \label{eq:reg_lossy_multi_prod}
\end{align}
where $\tilde{t}_{i,\vc{m}}\coloneqq\frac{1}{\sqrt{(\pi\sqrt{Kd})^{Kd}}}e^{-\frac{\sum_{k\in[K]}\|\mu_{x_i,k}-\mu_{y_i,k,m_k}^{(t)}\|^2}{\sqrt{Kd}}}$. The regularizer estimate is again $D_{\text{est}}=(D_{\text{var}}+D_{\text{prod}})/2$.

The optimal parameters are obtained by setting partial derivatives of $D_{\text{est}}$ to zero. Defining
\begin{align}
    \beta_{i,c,\vc{m}}&=\begin{cases}\frac{\eta_{i,\vc{m}}}{\sum_{\vc{m}'}\eta_{i,\vc{m}'}},&\text{if }c=y_i,\\0,&\text{otherwise},\end{cases} \qquad \eta_{i,\vc{m}}\coloneqq\alpha_{y_i,\vc{m}}^{(t)}e^{-\frac{\sum_{k\in[K]}\|\mu_{x_i,k}-\mu_{y_i,k,m_k}^{(t)}\|^2}{\sqrt{Kd}}},
\end{align}
with marginals $\gamma_{i,c,k,m}$, $\beta_{i,c,k,m}$ defined as before, $\tilde{\gamma}_{i,c,\vc{m}}=(\gamma_{i,c,\vc{m}}+\beta_{i,c,\vc{m}})/2$, and $\hat{\gamma}_{i,c,k,m}=(2\gamma_{i,c,k,m}+\beta_{i,c,k,m})/3$, the solutions are
\begin{align}
    \mu_{c,k,m}^*&=\frac{1}{\hat{b}_{c,k,m}}\sum_{i\in[b]}\hat{\gamma}_{i,c,k,m}\,\mu_{x_i,k}, \qquad {\sigma_{c,k,m,j}^*}^2=\frac{1}{b_{c,k,m}}\sum_{i\in[b]}\gamma_{i,c,k,m}\,\sigma_{x_i,k,j}^2,\nonumber\\
    \alpha_{c,\vc{m}}^*&=\tilde{b}_{c,\vc{m}}/\tilde{b}_c, \quad \tilde{b}_{c,\vc{m}}=\sum_{i\in[b]}\tilde{\gamma}_{i,c,\vc{m}}, \quad \tilde{b}_c=\sum_{\vc{m}}\tilde{b}_{c,\vc{m}},  \nonumber \\ \hat{b}_{c,k,m}&=\sum_{i\in[b]}\hat{\gamma}_{i,c,k,m},  b_{c,k,m}=\sum_{i\in[b]}\gamma_{i,c,k,m}. \label{eq:optimal_updates_lossy_multi}
\end{align}
As in the lossless case, $\mu_{c,k,m}^*$ and $\sigma_{c,k,m}^*$ depend only on the marginal coefficients and the $k$-th view's latent parameters, enabling the same distributed update protocol described in Appendix~\ref{sec:lossless_multi_view}. The parameters $\alpha_{c,\vc{m}}^{(t+1)}$, $\mu_{c,k,m}^{(t+1)}$, $\sigma_{c,k,m}^{(t+1)}$ are updated as moving averages of their current values and the above optimal solutions.

\textbf{Regularizer.} The estimate~\eqref{eq:reg_lossless_multi_estimate} simplifies to
\begin{align}
    \textnormal{Regularizer}(\vc{Q})=&-\sum_{i\in[b]}\log\left(\sum_{\vc{m}\in[M]^K}\alpha_{y_i,\vc{m}}^{(t)}e^{-\sum_{k\in[K]}D_{KL,\text{Lossy}}\big(P_{U_{i,k}|x_{i,k},\hat{w}_{e,k}}\|Q_{y_i,k,m_k}^{(t)}\big)}\right)\nonumber\\
    &-\frac{1}{2}\sum_{i\in[b]}\left(Kd\log\left(\pi e\sqrt{Kd}\right)+\log\Big(\sum_{\vc{m}}\alpha_{y_i,\vc{m}}^{(t)}\tilde{t}_{i,\vc{m}}\Big)\right). \label{eq:reg_lossy_multi_complete}
\end{align}


\section{Details of the experiments} \label{sec:details_exp}
This section provides additional details on the experiments that were conducted. The code used in the experiments is available at \url{https://github.com/PiotrKrasnowski/Gaussian_Product_Mixture_Priors_for_Multiview_Representation_Learning}. 

\subsection{Datasets}
\label{experiments:datasets}
In all experiments, we used the following image classification datasets:

\begin{itemize}[leftmargin=0.5cm]
    \item \textbf{CIFAR10} \cite{krizhevsky2009learning} - a dataset of 60,000 labeled images of dimension $32 \times 32 \times 3$ representing 10 different classes of animals and vehicles.

\item \textbf{CIFAR100} \cite{krizhevsky2009learning} - a dataset of 60,000 labeled images of dimension $32 \times 32 \times 3$ representing 100 different classes. 

\item \textbf{USPS} \cite{hull1994database}\footnote{\url{https://www.csie.ntu.edu.tw/~cjlin/libsvmtools/datasets/multiclass.html\#usps}} - a dataset of 9,298 labeled images of dimension $16 \times 16 \times 1$ representing 10 classes of handwritten digits.

\item \textbf{IMDB-WIKI}\footnote{https://data.vision.ee.ethz.ch/cvl/rrothe/imdb-wiki/} - a dataset of over 500k labeled images representing face images with gender and age labels (only normalized age was used in experiments).
\end{itemize}

All images were normalized before feeding them to the encoder. In the multiview scenario, each encoder received a duplicate of the same image such that each duplicate was independently corrupted by some distortion, as explained in Section~\ref{sec:experiments}.

\subsection{Architecture details}
\label{experiments:architecture}

The experiments were conducted using two types of encoder models: a custom convolutional encoder and a pre-trained ResNet18 followed by a linear layer (more specifically, the used model is the pre-trained model ``ResNet18\_Weights.IMAGENET1K\_V1'' in PyTorch for classification tasks or an untrained model with randomly initialized weights used for regression tasks). The architecture of the CNN-based encoder can be found in Table~\ref{tab::tab1}. This custom encoder is a concatenation of four convolutional layers and two linear layers. We apply max-pooling and a LeakyReLU activation function with a negative slope coefficient set to $0.1$. The encoders take uniformly re-scaled images as input and generate parameters $\mu_x$ and variance $\sigma_x^2$ of the latent variable of dimension $m=64$. Latent samples are produced using the reparameterization trick introduced by \cite{kingma2014auto}. Subsequently, the generated per-view latent samples are concatenated and fed into a decoder with a single linear layer and softmax activation function. The decoder's output is a soft class prediction for the classification task or the prediction of the continuous target value for the regression tasks.

Our tested encoders were complex enough to make them similar to ``a universal function approximator'', in line with \cite{dubois2020learning}. Conversely, we employ a straightforward decoder akin to \cite{alemi2016deep} to minimize the unwanted regularization caused by a highly complex decoder. This approach allows us to emphasize the advantages of our regularizer in terms of generalization performance.

\begin{table}[ht]
\centering
\caption{The architecture of the convolutional encoder used in the experiments. The convolutional layers are parameterized respectively by the number of input channels, the number of output channels, and the filter size. The linear layers are defined by their input and output sizes.\vspace{0.2 cm}}
\begin{tabular}{|cc|cc|cc|}
\hline
\multicolumn{2}{|c|}{Encoder}             & \multicolumn{2}{c|}{Encoder cont'd}        & \multicolumn{2}{c|}{Encoder cont'd}       \\ \hline
\multicolumn{1}{|c|}{Number} & Layer           & \multicolumn{1}{c|}{Number} & Layer             & \multicolumn{1}{c|}{Number} & Layer           \\ \hline
\multicolumn{1}{|c|}{1}  & Conv2D(3,8,5)  & \multicolumn{1}{c|}{6}  & Conv2D(16,16,3)  & \multicolumn{1}{c|}{11} & LeakyReLU(0.1)  \\ \hline
\multicolumn{1}{|c|}{2}  & Conv2D(8,8,5)  & \multicolumn{1}{c|}{7}  & LeakyReLU(0.1)   & \multicolumn{1}{c|}{12} & Linear(256,128) \\ \hline
\multicolumn{1}{|c|}{3}  & LeakyReLU(0.1) & \multicolumn{1}{c|}{8}  & MaxPool(2,2)     & \multicolumn{2}{c|}{Decoder}              \\ \hline
\multicolumn{1}{|c|}{4}  & MaxPool(2,2)   & \multicolumn{1}{c|}{9}  & Flatten          & \multicolumn{1}{c|}{1}  & Linear(64 $\cdot$ K,10)   \\ \hline
\multicolumn{1}{|c|}{5}  & Conv2D(8,16,3) & \multicolumn{1}{c|}{10} & Linear(1024,256) & \multicolumn{1}{c|}{2}  & Softmax         \\ \hline
\end{tabular}
\label{tab::tab1}
\end{table}

\subsection{Implementation and training details}
\label{experiments:implementation}

The PyTorch library \cite{paszke2019pytorch}, a GPU Tesla P100 with CUDA 11.0, and GPU NVIDIA H200 with CUDA 12.4 were utilized to train our prediction model. We employed the PyTorch Xavier initialization scheme \cite{glorot2010understanding} to initialize all weights, except biases set to zero. For optimization, we used the Adam optimizer \cite{KingmaB14} with parameters $\beta_1 = 0.5$ and $\beta_2 = 0.999$, an initial learning rate of $10^{-4}$, an exponential decay of 0.97, and a batch size of 128. 

In the training phase, we conducted a joint training of the models for 200 epochs with either the standard VIB or our Gaussian mixture objective functions. The Gaussian mixture priors were initialized using the approaches in~[Appendix~C.1]\cite{sefidgaran2025generalization} and~\ref{sec:initialization_multi_view}. Following the approach outlined in \cite{alemi2016deep}, we generated one latent sample per image during training and 5 samples during testing.

\subsection{Additional experiment results}
\label{experiments:additional_results}

Table \ref{table:experiments_multiview_additional} presents additional experiment results of the classification and regression tasks with the number of views $K$ equal to 2, 3, and 4. The description of different noise levels is shown in Table~\ref{table:noise_levels}.

\begin{table}[h]
 \renewcommand{\arraystretch}{1.5}
    \centering
    \begin{tabular}{|c|c|c|c|c|}
\hline
Level & Pixel Erasure Rate & Rotation $^{\circ}$ & Image Scale & Translations\\
\hline
Light & $5 \%$ & $[-5,5]$ &  $[0.9,1.1]$ & $(0\%,0\%) $ \\
\hline
Medium & $10 \%$& $[-7.5,7.5]$ &  $[0.8,1.2]$ & $(0\%,0\%) $  \\
\hline
Heavy & $20 \%$ & $[-10,10]$ &  $[0.6,1.4]$ & $(20\%,20\%)$ \\ \hline
Ultimate & $40 \%$ & $[-20,20]$ &  $[0.5,1.5]$ & $(40\%,40\%)$ \\ \hline
   \end{tabular}
    \caption{Different random distortion levels: values refer to the maximum thresholds and
the last column includes the maximum vertical and horizontal translations’ thresholds.}
    \label{table:noise_levels}
\end{table}

\begin{table}[h] \renewcommand{\arraystretch}{1.5}
\hspace{-.3 cm}    
   \begin{tabular}{|c|c|l|c|c|c|}
           \hline 
           \# & K & \hspace{3.8 cm} Scenario &  no reg. & VIB & \textbf{GPM-MDL} \\ \hline
           1 &2& CIFAR10, Enc.=CNN4, Dist.=(Light,Heavy) &  59.6\% & 59.7\% & \textbf{62.1\%}\\\hline
        2 &2& USPS, Enc.=CNN4, Dist.=(Light,Light) &  95.2\% & 95.3\% & \textbf{95.7\%}\\
            \hline
            3 & 2& CIFAR10, Enc.=CNN4, Dist.= Occ.(L,R) + (Medium, Medium) &  54.8\% & 55.3\% & \textbf{57.7\%}\\
           \hline
           4 & 2&  USPS, Enc.=CNN4, Dist.=Occ.(L,R) + (Heavy, Heavy) &  50.7\% & 51.5\% & \textbf{62.7\%}\\
           \hline
                5 & 3 & CIFAR100, Enc.=ResNet18, Dist.=(Light, Heavy, Heavy) &  37.5\% & 38.0\% & \textbf{42.7\%}\\
           \hline
           6 & 3 & CIFAR100, Enc.=ResNet18, Dist.=(Medium, Heavy, Heavy) &  32.4\% & 32.5\% & \textbf{36.6\%}\\
            \hline
             7 & 4 & CIFAR10, Enc.=CNN4, Dist.=(Light,Light,Light,Light) &  62.3\% & 62.8\% & \textbf{65.7\%}\\
           \hline
        8 & 4 & CIFAR10, Enc.=CNN4, Dist.=(Medium, Medium, Medium, Medium) &  60.2\% & 60.5\% & \textbf{63.9\%}\\
           \hline
            9 & 4 & CIFAR10, Enc.=CNN4, Dist.=(Medium, Heavy, Heavy, Heavy) & 57.4\% & 57.6\% & \textbf{60.0\%}\\\hline
           10 & 4 & USPS, Enc.=CNN4, Dist.= (Heavy, Heavy, Heavy, Heavy) & 58.7\% & 58.8\% &  \textbf{69.6\%}\\
           \hline
           11 & 4 & CIFAR10, Enc.=CNN4, Dist.=Occ.(LU,RU,LB,RB)& 62.0\% & 62.1\% & \textbf{64.6\%}\\
           \hline
           12 & 4 & CIFAR10, Enc.=CNN4, Dist.=Occ.(LU,RU,LB,RB)+(Light,\ldots,Light) & 58.5\% & 59.0\% & \textbf{62.0\%}\\ \hline
           13 & 8 & CIFAR10, Enc.=CNN4, Dist.=(Heavy,Heavy,Ultimate\ldots,Ultimate) & 25.6\% & 30.2\% & \textbf{33.5\%}\\ \hline
           \hline 14 & 2 & IMDB-WIKI, Enc.=ResNet18,  Dist.=  Occ.(L,R) 
 + (Light,Light) & 0.578 & 0.578 & \textbf{0.575}\\ 
            \hline 15 & 2 & IMDB-WIKI, Enc.=ResNet18, Dist.=Occ.(L,R)+(Medium,Medium) & 0.604 & 0.604 & \textbf{0.602}\\
        \hline
        \end{tabular}
    \caption{Test performance of Multiview representation learning models with different encoder architectures, and trained on selected datasets using no regularizer, per-view VIB \cite{alemi2016deep}, and our proposed Gaussian-product Mixture MDL (GPM-MDL). For classification, the test accuracy is expressed in [\%] of correct predictions. For regression, the accuracy is the mean squared prediction error of normalized age labels. The description of distortion levels is in Table~\ref{table:noise_levels}. Encoder, distortion, and occlusion are abbreviated as ``Enc.'', ``Dist.'', and ``Occ.'', respectively.}
    \label{table:experiments_multiview_additional}
\end{table}

\end{document}